\documentclass[10pt,journal,compsoc]{IEEEtran}
\ifCLASSOPTIONcompsoc
  \usepackage[nocompress]{cite}
\else
  \usepackage{cite}
\fi
\ifCLASSINFOpdf
\else
\fi
\usepackage{amssymb}
\usepackage{amsmath}
\usepackage{graphicx}
\usepackage{xspace}
\newcommand{\ie}{{\emph{i.e.}},\xspace}

\newcommand{\eg}{{\emph{e.g.}},\xspace}

\newcommand{\etal}{{\emph{et al.}}}
\newcommand{\aka}{{\emph{a.k.a.}},\xspace}
\usepackage{multirow}
\usepackage{pifont}
\usepackage{makecell}
\usepackage{color}
\usepackage{dsfont}

\newcommand{\vv}[1]{\mathbf{#1}}

\usepackage{subcaption}
\usepackage{engord}
\usepackage{booktabs} 

\begin{document}
%
\title{EvLight++: Low-Light Video Enhancement with an Event Camera: A Large-Scale Real-World Dataset, Novel Method, and More}

\author{Kanghao Chen$^*$, Guoqiang Liang$^*$, Hangyu Li, Yunfan Lu and Lin Wang$^\dagger$,~\IEEEmembership{Member,~IEEE,}
\thanks{$^*$ Equal contribution}
\thanks{$^\dagger$ Corresponding author}
\thanks{Kanghao Chen, Guoqiang Liang, Hangyu Li, Yunfan Lu are with AI Thrust, Information Hub, Hong Kong University of Science and Technology (Guangzhou), China}
\thanks{Lin Wang is with AI Thrust, HKUST-GZ and Dept.
of CSE, HKUST, China}

}

%
%

\markboth{Journal of \LaTeX\ Class Files,~Vol.~14, No.~8, August~2015}%
{Shell \MakeLowercase{\textit{et al.}}: Bare Demo of IEEEtran.cls for Computer Society Journals}
%



\IEEEtitleabstractindextext{%
\begin{abstract}
Event cameras offer significant advantages for low-light video enhancement, primarily due to their high dynamic range. Current research, however, is severely limited by the absence of large-scale, real-world, and spatio-temporally aligned event-video datasets. To address this, we introduce a large-scale dataset with over \textbf{30,000} pairs of frames and events captured under varying illumination.  This dataset was curated using a robotic arm that traces a consistent \textbf{non-linear} trajectory, achieving spatial alignment precision under \textbf{0.03mm} and temporal alignment with errors under \textbf{0.01s} for 90\% of the dataset. Based on the dataset, we propose \textbf{EvLight++}, a novel event-guided low-light video enhancement approach designed for robust performance in real-world scenarios. Firstly, we design a multi-scale holistic fusion branch to integrate structural and textural information from both images and events. To counteract variations in regional illumination and noise, we introduce Signal-to-Noise Ratio (SNR)-guided regional feature selection, enhancing features from high SNR regions and augmenting those from low SNR regions by extracting structural information from events. To incorporate temporal information and ensure temporal coherence, we further introduce a recurrent module and temporal loss in the whole pipeline. Extensive experiments on our and the synthetic SDSD dataset demonstrate that EvLight++ significantly outperforms both single image- and video-based methods by \textbf{1.37} dB and \textbf{3.71} dB, respectively.  To further explore its potential in downstream tasks like semantic segmentation and monocular depth estimation, we extend our datasets by adding pseudo segmentation and depth labels via meticulous annotation efforts with foundation models. Experiments under diverse low-light scenes show that the enhanced results achieve a 15.97\% improvement in mIoU for semantic segmentation.

\end{abstract}

}

\maketitle

\IEEEdisplaynontitleabstractindextext

%
\IEEEpeerreviewmaketitle

\IEEEraisesectionheading{\section{Introduction}\label{sec:introduction}}

%
%
%
%
\begin{figure*}[t!]
\centering
\includegraphics[width=1\linewidth]{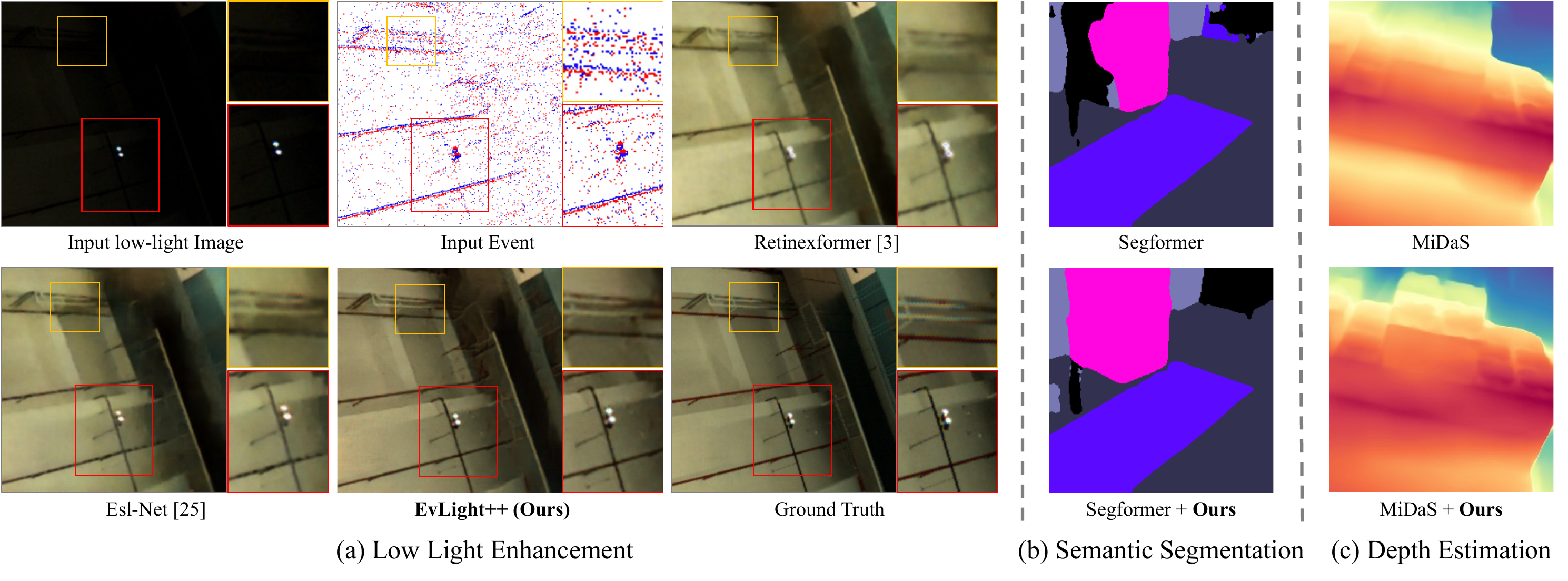}
\vspace{-10pt}
    \caption{(a) A challenging example of our dataset containing an extremely low-light image and events. Compared with the result from a SOTA frame-based method Retinexformer~\cite{cai2023retinexformer}, our EvLight++ not only recovers the structure details (\eg the pipe on the ceiling) but also avoids over-enhancement and saturation in the bright regions (\eg the lights). (b\&c) Based on the enhanced outputs, semantic segmentation, and depth estimation are conducted with the off-the-shelf models. }
    \label{fig:teaserfigure}
\centering
\vspace{-10pt}
\end{figure*}

\IEEEPARstart{I}{mages} captured under sub-optimal lighting conditions typically suffer from various types of degradation, including poor visibility, noise, and color inaccuracies~\cite{li2021low}. Consequently, low-light enhancement (LLE) for videos is a critical process for improving image quality in such conditions. LLE plays a vital role in enabling more accurate performance of downstream tasks, such as face detection~\cite{ma2022toward, yu2021single}, nighttime semantic segmentation~\cite{pan2023towards}, and depth estimation~\cite{pan2024srfnet}.
With the advent of deep learning, a plethora of frame-based methods have been developed to enhance contrast~\cite{zhang2021beyond12}, reduce noise~\cite{xu2020learning}, and correct color imbalances~\cite{wang2019underexposed}. Despite advanced performance, they often struggle with issues like unbalanced exposure and color distortion, especially when visual details such as edges captured by frame-based cameras are not distinct, as illustrated in Fig.~\ref{fig:teaserfigure} (a).

Event cameras, \aka bio-inspired sensors, produce event streams characterized by high dynamic range (HDR) and high temporal resolution~\cite{scheerlinck2019ced,zheng2023deep}. Despite their potential, few studies have explored integrating frame-based and event cameras to tackle LLE challenges~\cite{zhang2020learning,liu2023low,jiang2023event,liang2023coherent}. A significant obstacle remains the scarce availability of large-scale, real-world datasets with spatially and temporally aligned images and events. For instance, ~\cite{zhang2020learning} introduces an unsupervised framework that does not require paired event-image data, while~\cite{liu2023low,liang2023coherent} utilize synthetic datasets for training. Nevertheless, these approaches fall short in real-world low-light conditions. The LLE dataset~\cite{jiang2023event} provides a real-world event-image dataset with paired low-light/normal-light sequences, obtained through simple manipulations like adjusting indoor lighting or outdoor exposure time, keeping the camera stationary. However, similar to the earlier frame-based SMID dataset~\cite{chen2019seeing22}, it is restricted to static scenes only.
Recently, LLE-VOS~\cite{li2024event} has recorded paired low-light images/events with corresponding normal-light images/events using an optical system incorporating a beam splitter. 
However, LLE-VOS is designed for low-light video object segmentation and is limited in scene diversity, featuring only five different indoor scenes and outdoor videos recorded exclusively during the daytime, which is insufficient for training the LLE task.

In this paper, we introduce the \textbf{SDE} dataset, a comprehensive real-world dataset containing over 30K pairs of spatio-temporally aligned video frames and events captured under both low-light and normal-light conditions (see Sec.~\ref{sec:dataset} and Fig.~\ref{fig:data_sample} (a)). Constructing such a dataset presents inherent challenges due to the complexities of ensuring precise spatial and temporal alignment between paired sequences, particularly in dynamic scenes with nonlinear motion.
To this end, we developed a robotic alignment system employing a DAVIS346 event camera mounted on a Universal UR5 robotic arm (Fig.~\ref{fig:mechanism}). This system achieves remarkable spatial accuracy with an error margin of merely \textbf{0.03mm}, significantly surpassing the 1mm error of the frame-based SDSD dataset~\cite{wang2021seeing21}. Moreover, it handles non-linear motions with complex trajectories, greatly enhancing the diversity and applicability of our dataset to real-world scenarios compared to the uniform linear motion in SDSD and the static scenes in the LLE dataset~\cite{jiang2023event}.
For temporal alignment, traditional methods of clipping sequences by motion start and end timestamps often introduce random errors due to varying intervals ({\textit{blue}} regions in Fig.~\ref{fig:dataset-temporal-alignment} (a)) between the motion start timestamps ({left \textit{orange} line}) and the timestamps of the initial frame ({\textit{black} line}). To mitigate these discrepancies, we propose a novel matching alignment strategy. 

Moreover, we facilitate our dataset to benchmark downstream tasks by providing pseudo semantic and depth estimation labels generated through the vision foundation models, as shown in Fig.~\ref{fig:data_sample} (b). Given the precise spatial and temporal alignment between the low-light and normal-light sequences, it is feasible to transfer the pseudo labels from the normal-light sequences to the low-light sequences, as the pseudo labels on the normal-light data are more accurate and reliable. Specifically, for semantic segmentation, we utilize SSA~\cite{chen2023semantic}, an advanced tool built on top of SAM to predict semantic categories for each mask. For depth estimation, we leverage the monocular depth estimation foundation model, Depth Anything V2~\cite{depth_anything_v2}. With these pseudo labels and the paired frames and events, our dataset further enables comprehensive evaluation of downstream tasks for enhanced results.

Buttressed by our dataset, we introduce \textbf{EvLight++}, an event-guided LLE method designed for robust performance in real-world low-light conditions. The fundamental concept is that while low-light images provide critical color information, events capture essential edge details. However, each modality can exhibit distinct noise patterns, and directly fusing these features, as commonly practiced (\eg in~\cite{jiang2023event}), may exacerbate noise issues in different areas of the inputs, as illustrated in the blue box in Fig.~\ref{fig:visual-sde} (g).

To tackle these problems, \textit{our key idea is to fuse event and image features holistically, followed by a selective region-wise manner to extract the textural and structural information with the guidance of Signal-to-Noise-Ratio (SNR) prior information}. 
To ensure robustness against variations in the regional illumination and noise, we propose an SNR-guided feature selection strategy, extracting image features from high SNR regions and event features from low SNR areas, thus preserving critical textural and structural details (Sec.~\ref{sec:regional}).
Furthermore, we employ an attention-based holistic fusion branch to broadly integrate structural and textural information from both events and images (Sec.~\ref{sec:holistic}). This process involves a fusion block with channel attention, merging holistic and regional features from both modalities. Unlike existing methods that focus solely on single-frame enhancement without accounting for temporal dynamics, our framework further incorporates a recursive module and temporal loss to leverage temporal information more effectively.


To evaluate the effectiveness of our dataset and the proposed approach,
We conduct extensive experiments comparing our EvLight++ with existing frame-based methods, such as RetinexFormer~\cite{cai2023retinexformer}, and event-guided methods, such as~\cite{liu2023low}, on both our real-world SDE and the synthetic SDSD dataset~\cite{wang2021seeing21}, which includes events generated via an event simulator~\cite{hu2021v2e}. The results demonstrate that EvLight++ effectively enhances a wide range of underexposed images in extremely low-light conditions, as shown in Fig.~\ref{fig:teaserfigure} (b\&c).

\begin{table*}[!]
    \centering
    \caption{A summary of existing real-world image-event datasets. Note that images in DVS-Dark are gray-scale.}
    \setlength{\tabcolsep}{2pt}
    \resizebox{1\linewidth}{!}{
\begin{tabular}{lccccccc}
\hline
Dataset  & Release & Dynamic Scene & Data Description  & Frames  & Devices & Labels for Downstream Tasks \\ \hline
DVS-Dark \cite{zhang2020learning}  & \ding{55} & \ding{51} & Low-light images and pair events streams  & 17,765  & DAVIS240C event camera & \ding{55}\\ 
LIE \cite{jiang2023event}  & \ding{55} & \ding{55} & Low-light/normal-light images and pair events streams  & 2,231  & DAVIS346 event camera & \ding{55}\\ 
EvLowLight \cite{liang2023coherent}  & \ding{55} & \ding{51} & Low-light images and pair events streams  & ---  & {\begin{tabular}[c]{@{}c@{}}DAVIS346 event camera +\\FLIR Chameleon 3 Color RGB camera \end{tabular}} & \ding{55}\\ 
LLE-VOS \cite{li2024event}  & \ding{51} & \ding{51} & Low-light/normal-light images and pair events streams  & 5,600 &  2$\times$ DAVIS346 event cameras & Object Mask\\
RLED \cite{liu2024seeing}  & \ding{55} & \ding{51} & Normal-light images and pair low-light events streams  & 64,200  &{\begin{tabular}[c]{@{}c@{}}Prophesee EVK4 event camera +\\FLIR BFS-U3-32S4C RGB camera \end{tabular}}  &\ding{55}\\
\textbf{SDE (Ours)}  & \ding{51} & \ding{51} & Low-light/normal-light images and pair events streams  & 31,477  & DAVIS346 event camera & Semantic Mask\&Depth Map\\ \hline
\end{tabular}
}

    \label{tab:dataset_comparison}
    \vspace{-8pt}
\end{table*}

This manuscript presents a substantial improvement over our CVPR 2024 work~\cite{liang2024towards}, achieved by extending the dataset, method, experiments, and downstream tasks in three major aspects:
\begin{enumerate}
    \item \textbf{Enhanced Dataset for Downstream Tasks:} We have enhanced the dataset to support downstream tasks, \ie semantic segmentation and monocular depth estimation, by providing pseudo labels generated using recent large foundation models. Specifically: 
    \begin{itemize}
        \item  For semantic segmentation, we utilize SSA~\cite{chen2023semantic} to generate semantic labels for each mask, as detailed in Sec.~\ref{sec:annotation}. Additionally, we manually define the semantic classes based on those in ADE-20K~\cite{zhou2019semantic} and filter the samples with well-annotated labels.
        \item For depth estimation, we employ the large-scale Depth Anything V2~\cite{depth_anything_v2} model to produce depth maps, as discussed in Sec.~\ref{sec:annotation}.   To ensure the robustness of the annotation pipelines, we manually rotate each sequence to a standard upright perspective, addressing the model's sensitivity to rotation. 
    \end{itemize} 
 This process produces high-quality labels for the paired low-light and normal-light sequences, as shown in Fig.~\ref{fig:data_sample} (b). The new dataset with pseudo labels provides a benchmark for evaluating our method tailored for these downstream tasks. Extensive discussions underscore the dataset's essential role in event-based LLE tasks.

\item \textbf{Temporal Information for Video Enhancement:} 
Previous methods that rely solely on single frames and corresponding events are prone to issues like illumination flicker. To address this, we propose incorporating temporal cues and constraints to further enhance the performance.
\begin{itemize}
    \item To facilitate video enhancement and leverage temporal information, we incorporate a recurrent network to recover illumination-consistent normal-light videos from low-light videos and event data. For convenience, we employ ConvGRU~\cite{ballas2015delving} to model the long-range temporal dependencies by enhancing the holistic feature (Sec.~\ref{sec:recurrent}). 
\item Moreover, a temporal loss is introduced to impose the enhancement fidelity of temporal consistency (Sec.~\ref{sec:loss}).
\end{itemize}

    \item \textbf{Practical Applications and Comprehensive Analysis:} Our dataset and approach pave the way for practical applications of the LLE task. 
    \begin{itemize}
    \item  The effectiveness of our model has been tested through downstream tasks of semantic segmentation and depth estimation (see Sec.~\ref{sec:downstream_result}). The quantitative results presented in Tab.~\ref{tab:downstream_seg} and Tab.~\ref{tab:downstream_depth} offer a comprehensive analysis of its potential applications. The qualitative results in Fig.~\ref{fig:downstream_seg} and Fig.~\ref{fig:downstream_depth} further demonstrate the superiority of our method. 
    \item  We conduct extensive experiments and ablation studies on our video pipeline with temporal enhancement to validate the effectiveness of the introduced modules. Our in-depth exploration of the event-based LLE dataset and framework design provides valuable insights for future research.
    \end{itemize}
  
\end{enumerate}

\section{Related Work}


\noindent\textbf{Low-light Enhancement Datasets. }
The performance of LIE heavily relies on the quality of the training datasets~\cite{fu2023dancing} for either images~\cite{wei2018deep,chen2018learning,bychkovsky2011learning} or videos~\cite{lee2023humanpose, wang2019ehsc, chen2019seeing22,wang2021seeing21,fu2023dancing,jiang2019smoid}. 
Compared to the image datasets, ensuring the quality of video datasets is challenging due to the difficulties in achieving spatial and temporal alignment between each input frame and the ground truth (GT).
SMID~\cite{jiang2019smoid} addresses these challenges by recording static scenes through the controlled exposure time. SDSD~\cite{wang2021seeing21} obtains pairs of videos under different lighting conditions with a camera mounted on an electric slide rail for two rounds, ensuring spatial alignment. 
Temporal alignment is then manually achieved by selecting the first moving frame. 
DID~\cite{fu2023dancing} ensures spatial and temporal alignment by capturing video datasets frame by frame using an electric gimbal.

In this paper, we focus on low-light event-video paired datasets. 
Tab.~\ref{tab:dataset_comparison} summarizes existing real-world image-event datasets, including those for low-light scenes. 
DVS-Dark~\cite{zhang2020learning} offers unpaired low-/normal-light images/events, facing similar alignment challenges to the video datasets. 
EvLowLight~\cite{liang2023coherent} contains only low-light images/events without corresponding normal-light images/events as GT, while RLED~\cite{liu2024seeing} focuses on event-to-video reconstruction, providing only low-light events and normal-light images. 
LIE~\cite{jiang2023event} is a real-world image-event dataset captured by adjusting the camera's light intake in static scenes, with events triggered by light changes (indoor) and exposure times (outdoor).
Recently, LLE-VOS~\cite{li2024event} has captured paired low-light and normal-light images and events for low-light video object segmentation using two DAVIS event cameras. 
LLE-VOS synchronizes these cameras using an optical system with a beam splitter to ensure a consistent viewpoint with a synchronous cable for simultaneous scene capture.
Regarding labels for downstream tasks, only LLE-VOS provides object masks in low-light conditions based on aligned normal-light frames. In contrast, our dataset provides semantic masks and depth labels, enabling convenient evaluation of the effectiveness of LLE methods for semantic segmentation and monocular depth estimation.

\noindent\textbf{Low-light image enhancement.}
Frame-based methods for low-light image enhancement can be divided into non-learning-based methods~\cite{arici2009histogram1,nakai2013color2,guo2016lime,fu2016weighted9,xu2020star10} and learning-based methods~\cite{wei2018deep,zhang2019kindling13,zhang2021beyond12,wu2022uretinex5,fu2023you6,cai2023retinexformer,wang2019underexposed,xu2022snr4,xu2023low7,wang2023low8,wu2023learning}.
Non-learning-based methods typically rely on handcrafted features, such as histogram equalization~\cite{arici2009histogram1,nakai2013color2} and the Retinex theory~\cite{guo2016lime,fu2016weighted9,xu2020star10}. 
Nonetheless, these methods lead to the absence of adaptivity and efficiency~\cite{wu2022uretinex5}.
With the development of deep learning, an increasing number of learning-based methods have emerged, which can be bifurcated as Retinex-based methods~\cite{wei2018deep,zhang2019kindling13,zhang2021beyond12,wu2022uretinex5,fu2023you6,cai2023retinexformer} and non-Retinex-based methods~\cite{wang2019underexposed,xu2022snr4,xu2023low7,wang2023low8,wu2023learning}.
For instance, SNR-Aware~\cite{xu2022snr4} uses Signal-to-Noise-Ratio-aware transformers and convolutional models for dynamic pixel enhancement.
However, these frame-based approaches often result in blurry outcomes and low Structural Similarity (SSIM) due to the buried edge in low-light images.

\noindent\textbf{Low-light video enhancement.}
Studying low-light video enactment is necessary since simply applying low-light image enactment methods will cause the flickering problem due to temporal inconsistency.
The first line of research is to apply the temporal consistency loss to relieve the flickering problem.
SMID~\cite{chen2019seeing22} collects short- and long-exposure video pairs of static scenes to ensure different short exposure frames close to the exposure frame with the temporal consistency loss.
Zhang \etal~\cite{zhang2021learning} and LVE-S2D~\cite{peng2022lve} synthesizing video sequences from the image captured in static scenes to extend the temporal consistency loss to dynamic scenes.
Another line of research is to extract the temporal alignment information from the multi-frame inputs to the desired frame. 
MBLLEN~\cite{lv2018mbllen} introduces 3D convolution layers to obtain temporal information from image sequences and SDSD~\cite{wang2021seeing21} proposes the progressive alignment to align the neighboring frames into the central one.
LAN~\cite{fu2023dancing} estimate the reflectance and illumination components of predicted normal light frames with the input of multi-frame, respectively.
However, due to the low dynamic range of conventional cameras and lack of inter-frame information, theses methods are easy to generate artifacts in extreme low-light or fast-moving scenes~\cite{liu2023low}.

\noindent\textbf{Event-guidied low-light enhancement.}
Zhang \etal~\cite{zhang2020learning} and NER-Net~\cite{liu2024seeing} focus on reconstructing grayscale images from low-light events but face challenges in preserving the details using only brightness changes from events.
Recently, some researchers have utilized events for guiding low-light image enhancement~\cite{jiang2023event, jin2023event} and low-light video enhancement~\cite{liu2023low,liang2023coherent}.
ELIE~\cite{jiang2023event} proposes a two-stage residual fusion module to integrate additional structural information of event features with image features thanks to HDR of event cameras.
EGLLIE~\cite{jin2023event} introduces an event-guided low-light enhancement method with an end-to-end dual branch GAN by using events as conditions for the discriminator network that can generate images with better textures.
For low-light video enhancement, Liu \etal~\cite{liu2023low} address multi-frame fusion artifacts while realizing temporal consistency. 
They synthesize events to record intensity and motion information between adjacent images by fusing these events with images to guide the enhancement of low-light videos.
EvLowLight~\cite{liang2023coherent} establishes temporal coherence by jointly estimating motion from events and frames while ensuring the spatial coherence between events and frames with different spatial resolutions.
However, these methods directly fuse features extracted from events and images without considering the discrepancy of the noise at the different local regions of events and images. Also, these methods model temporal dependencies by introducing flow estimation based on multi-frame input, which requires a large amount of GPU memory. In contrast, we incorporate a recurrent framework to alleviate the high memory cost while improving temporal modeling capabilities.

\vspace{-3pt}
\section{Mechanism System and Our SDE Dataset}
\label{sec:dataset}

\begin{figure}[t]
\centering
    \includegraphics[width=1\linewidth]{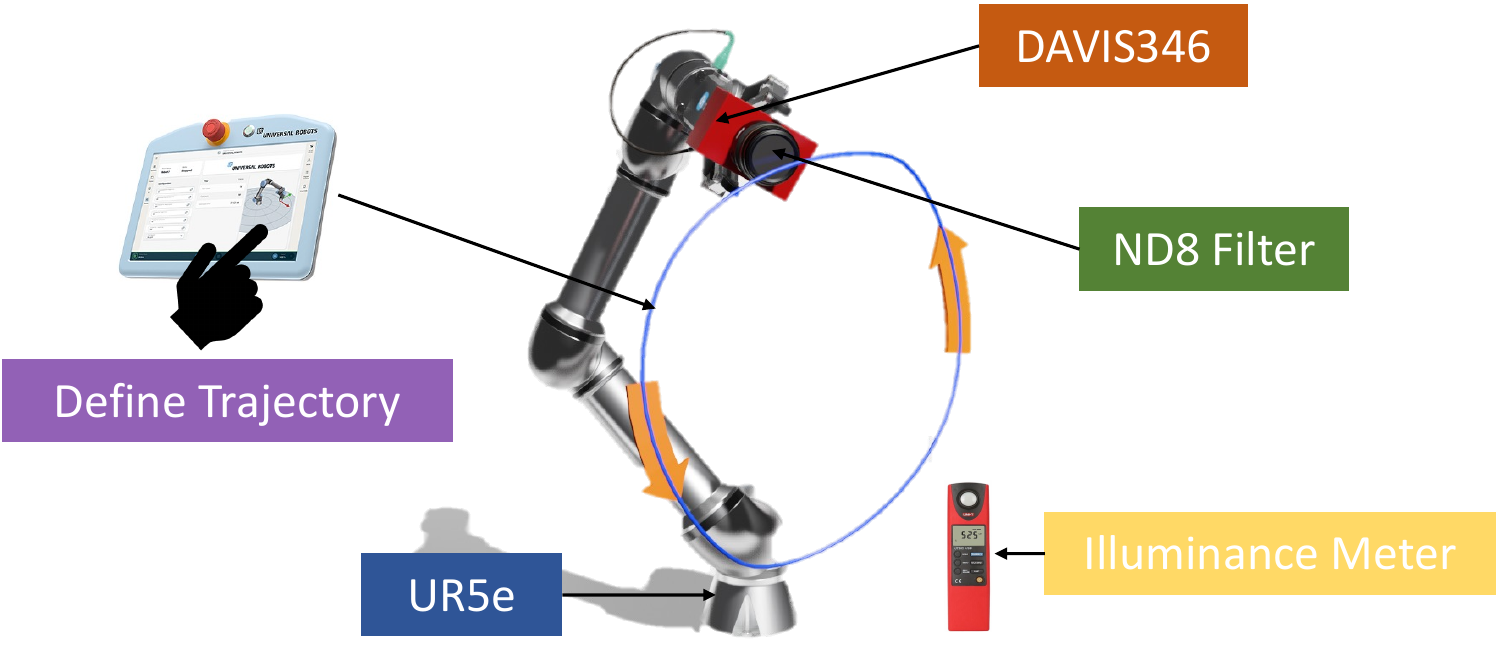}
    \caption{An illustration of our mechanism system to collect spatially aligned image-event dataset by mounting a DAVIS 346 event camera on the UR5e robotic arm and recording the sequences with the pre-defined trajectory. The corresponding low-light sequence is captured by applying an ND8 filter to the camera lens.}
    \label{fig:mechanism}
\centering
\vspace{-8pt}
\end{figure}

\begin{figure*}[t]
\centering
    \includegraphics[width=1\linewidth]{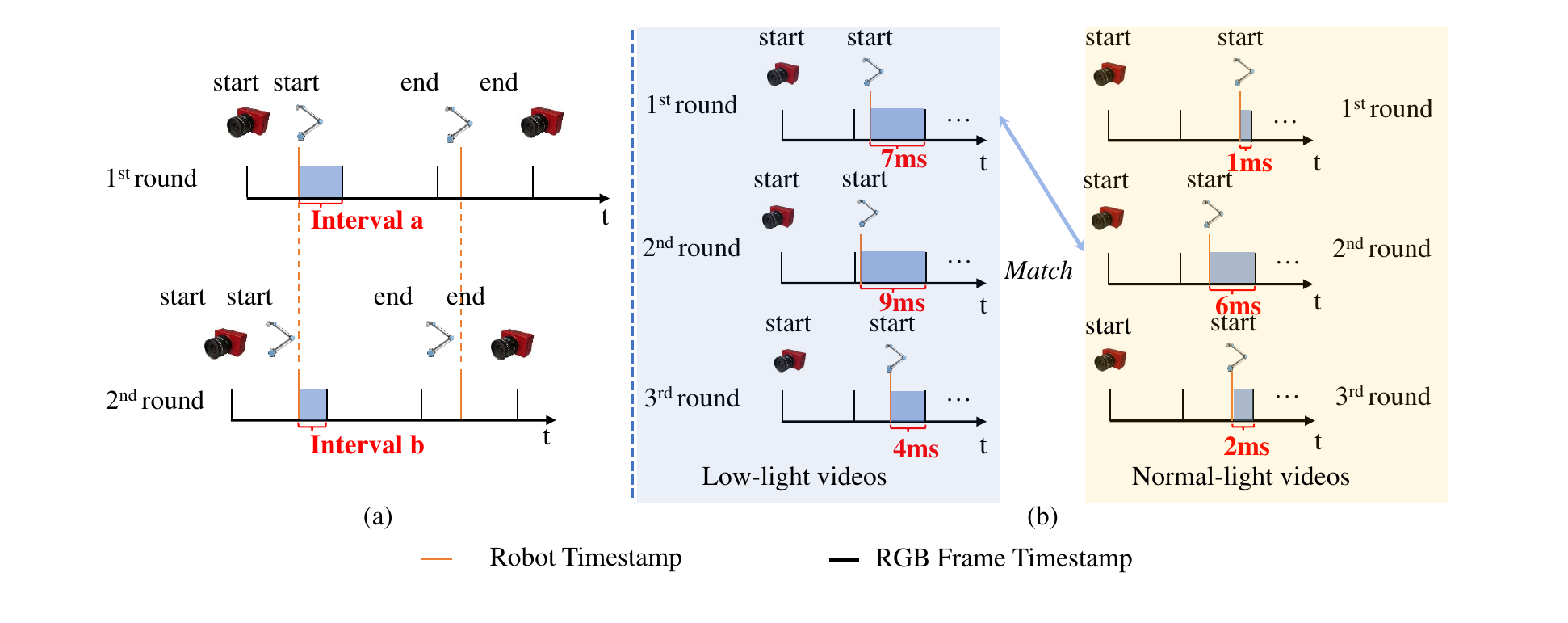}
    \caption{(a) An illustration of the variable time interval between the start timestamp of the trajectory and the first frame timestamp in each sequence. (b) An example of the matching alignment strategy.
    }
    \label{fig:dataset-temporal-alignment}
\centering
\end{figure*}

\begin{figure*}[t]
\centering
    \includegraphics[width=1\linewidth]{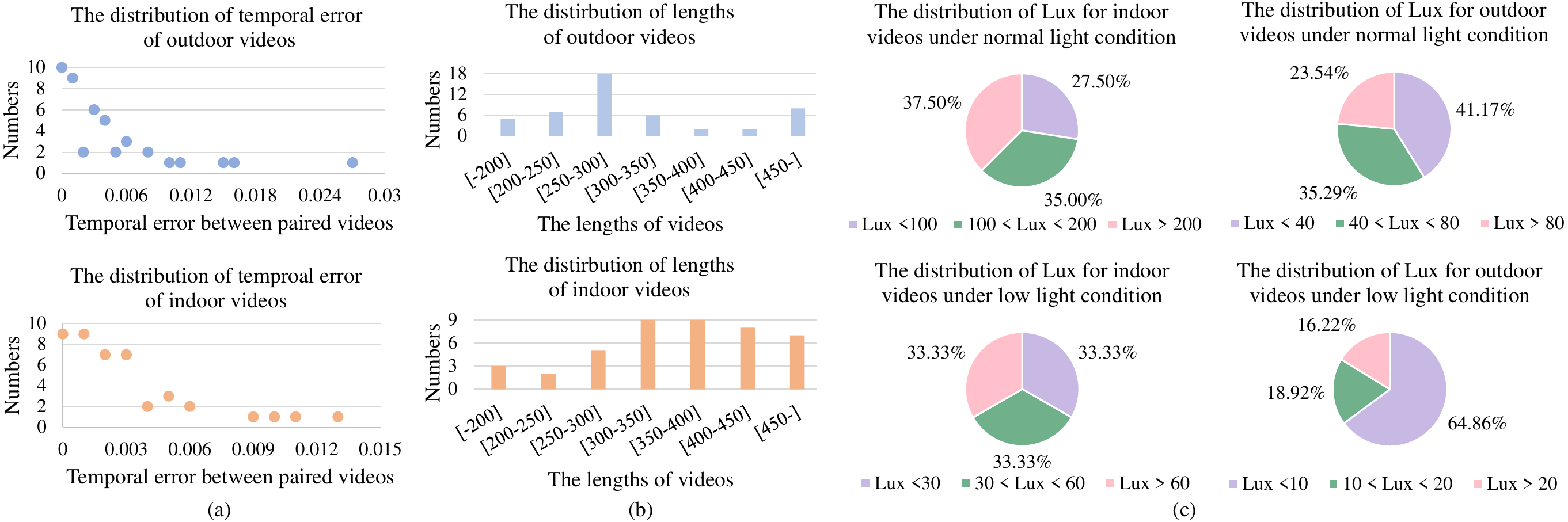}
    \caption{(a) Distribution of temporal alignment error (measured in seconds) of our dataset. (b) Distribution of video length of our dataset. (c) Illumination distribution in the filming environment.
    }
    \label{fig:dataset-distribution}
\centering
\end{figure*}

Capturing paired dynamic sequences from real-world scenes presents a formidable challenge, primarily attributed to the complexity involved in ensuring spatial and temporal alignment under varying illumination conditions.
The first line of approaches employs a stereo camera system to simultaneously record identical scenes, using non-linear transformations and cropping like DPED~\cite{ignatov2017dslr}.
However, under low-light visibility, this method encounters difficulties in computing and matching SIFT key points~\cite {lowe2004distinctive}, essential for determining and acquiring overlapping segments of synchronized videos.
The second line of approaches~\cite{jiang2019smoid,lee2023humanpose} utilizes an optical system with a beam splitter, enabling two cameras to capture the same view.
Nonetheless, achieving perfect alignment between these two cameras presents challenges, particularly when they encounter varying lighting conditions, which often leads to spatial misalignments, as noted in~\cite{lee2023humanpose,rim2020real,liang2023coherent}.
The third approach involves designing mechatronic systems to achieve precise spatial and temporal alignment. SDSD's~\cite{wang2021seeing21} system employs an electric slide rail to mount the camera, facilitating the separate capture of low-light and normal-light videos in two distinct sessions. Conversely, the system of DID~\cite{fu2023dancing} utilizes an electric gimbal to incrementally move the camera, allowing for the sequential capture of paired videos frame by frame. However, these systems are limited by the linear motion of the slide rail and the 2D rotation of the gimbal. 
In contrast, our design incorporates a robotic alignment system equipped with an event camera, enabling the capture of paired RGB images and events under varied lighting conditions. 
This system uniquely supports non-linear movements along complex trajectories.

\vspace{-10pt}
\subsection{Overview of Robotic System}
To ensure precise spatial-temporal alignment, we design a robotic alignment system.
It comprises several key components: a DAVIS346 event camera~\cite{taverni2018front}, a robotic arm Universal Robot UR5e, an ND 8 filter from NiSi, and an illuminance meter, as shown in Fig.~\ref{fig:mechanism}.
The specific settings for the robotic arm and event camera are contingent upon predefined paths and the lighting conditions within the scenes. 
In practical terms, the camera's exposure time varies between 10 $ms$ and 80 $ms$, maintaining a fixed frame interval of 5 $ms$ \footnote{DV software may adjust the interval between frame and exposure to suit the desired exposure duration.}. 

\subsection{Spatial and Temporal Alignments Pipeline}

\subsubsection{Data Capture with Spatial Alignment}
To ensure the spatial alignment of paired sequences, a robotic arm (Universal UR5), exhibiting a minimal repeated error of 0.03mm, is equipped to capture sequences following an identical trajectory. 
We set the robotic system with a pre-defined trajectory and a DAVIS 346 event camera with fixed parameters, \eg exposure time.
Firstly, paired image and event sequences are acquired under normal lighting conditions.
Subsequently, an ND8 filter is applied to the camera lens, which facilitates the capture of low-light sequences while maintaining consistent camera parameters, such as exposure time and frame intervals.


\begin{figure*}[t]
\centering
    \includegraphics[width=1\linewidth]{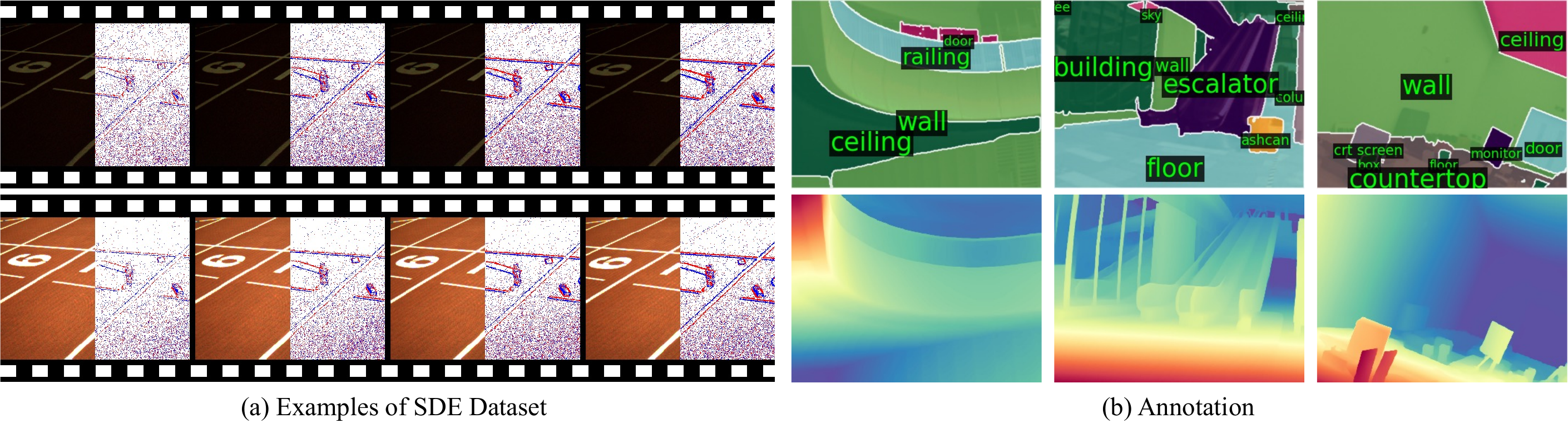}
    \vspace{-8pt}
    \caption{(a) An example of our dataset with images and paired events captured in low-light (with an ND8 filter) and normal-light conditions. (b) Examples of the downstream annotations of semantic segmentation and depth estimation.}
    \vspace{-10pt}
    \label{fig:data_sample}
\centering
\end{figure*}

\subsubsection{Temporal Alignment of Low-light/Normal-light sequences}
The alignment of SDSD~\cite{wang2021seeing21} dataset involves a manual selection of the initial and final frames of each paired video, based on the motion states depicted in the videos, leading to inevitable bias.
To mitigate this problem, initial temporal alignment is performed by trimming the sequences based on the start and end timestamps of a pre-defined trajectory.
However, even with consistent settings for exposure time and frame intervals, there exists a variable time interval between the start timestamp of the trajectory and the first frame timestamp captured post-initiation of the trajectory in each sequence.
The bias causes the misalignment between each low-light sequence and its normal-light sequence pair, particularly in complex motion paths.

To achieve further alignment, we introduce a matching alignment strategy, wherein sequences from each scene are captured multiple times to minimize the alignment error to the largest extent, as shown in Fig.~\ref{fig:dataset-temporal-alignment}.
Fig.~\ref{fig:dataset-temporal-alignment} (a) illustrates the variance between intervals a and b ({highlighted in \textit{blue} regions}), despite identical camera and robot settings, leading to unpredictable temporal alignment errors.
Our approach to mitigating this issue involves the introduction of a novel matching alignment strategy.
In particular, we capture 6 paired event-image sequences per scene —three in low-light and three in normal-light conditionals. 
These 6 sequences are trimmed to the predefined trajectory's start and end timestamps, ensuring uniform content across all videos.
Subsequently, the time intervals between the trajectory's start timestamps and the initial frame timestamps of each trimmed sequence are calculated.
As shown in Fig.~\ref{fig:dataset-temporal-alignment} (b), the time intervals of 6 sequences are different, and we match the low-light sequence with the normal-light sequence, which has the minimal absolute errors of their time intervals; thus, we can reduce the misalignment caused by the random time interval. For example, we perform matching between the \engordnumber{1} low-light sequence and the \engordnumber{2} normal-light sequence, achieving minimal absolute errors of $1ms$.
Using our matching alignment strategy, 90\% of the datasets achieve temporal alignment errors under 0.01 seconds, with maximum errors of \textbf{0.013s} indoors and \textbf{0.027s} outdoors.

\subsection{The proposed SDE dataset}
\subsubsection{Paired Low-light/Normal-light Dataset}
Based on the video capture and alignment pipeline, we have collected 91 paired low-light/normal-light sequences with videos and the corresponding event data.
Following SDSD dataset~\cite{wang2021seeing21}, we distribute our dataset into outdoor (43 sequences) and indoor (48 sequences) according to the scenarios. Each video consists of 145~888 frames and the corresponding events, and the resolution is 346$\times$260. Fig.~\ref{fig:dataset-distribution} visualizes the temporal alignment error distribution (measured in seconds), the video length distribution within our dataset, and the illumination variations across filming environments. 
Exceeding the measured Lux levels of the SDSD dataset, we capture sequences in better-illuminated environments due to the limitations imposed by the lens (with an aperture of f2.0) and sensor size. Fig.~\ref{fig:data_sample} (a) provides an example of a sequence under low-light and normal-light conditions, along with the corresponding event frames.

\subsubsection{Labeling for Downstream Tasks}
\label{sec:annotation}
The Low-Light Enhancement (LLE) task is essential for downstream tasks, as most existing downstream methods are optimized for normal-light data and often struggle with low-light scenarios. To demonstrate the generalization and practical value of LLE methods, it is important to evaluate their performance on downstream tasks. While almost no existing LLE dataset includes pseudo labels for downstream tasks (except for LLE-VOS~\cite{li2024event}), our SDE dataset provides high-quality and abundant downstream annotations. Given the precise alignment of paired normal-light and low-light sequences, we can label the normal-light sequences and transfer these annotations directly to the low-light sequences.
We focus on two key downstream vision tasks: semantic segmentation and monocular depth estimation.

\noindent\textbf{Semantic Segmentation:} We use the Semantic Segmentation Anything (SSA)~\cite{chen2023semantic} pipeline  to enrich our dataset with semantic category annotations. 
This model leverages SAM~\cite{kirillov2023segment}, a large-scale model for arbitrary object segmentation. 
While SAM excels in mask generation, it does not assign semantic categories to these masks. 
By combining SAM with traditional semantic segmentation models, SSA efficiently assigns categories to masks and generates precise segmentation masks, creating an effective annotation system.
Specifically, we use the pre-trained OneFormer~\cite{jain2023oneformer} on the ADE-20k~\cite{zhou2019semantic} dataset as the semantic branch. 
To tailor the annotations to our dataset, we define a subset of categories based on those in ADE-20k. 
Additionally, we observed that SSA is sensitive to camera rotation. 
Therefore, we manually annotate the rotation angle for each sequence to achieve more precise results. 
Fig.~\ref{fig:data_sample} (b, first row) shows examples of the annotation results, highlighting their high quality and robustness.

\begin{figure*}[t]
\vspace{-8pt}
\centering
    \includegraphics[width=.98\linewidth]{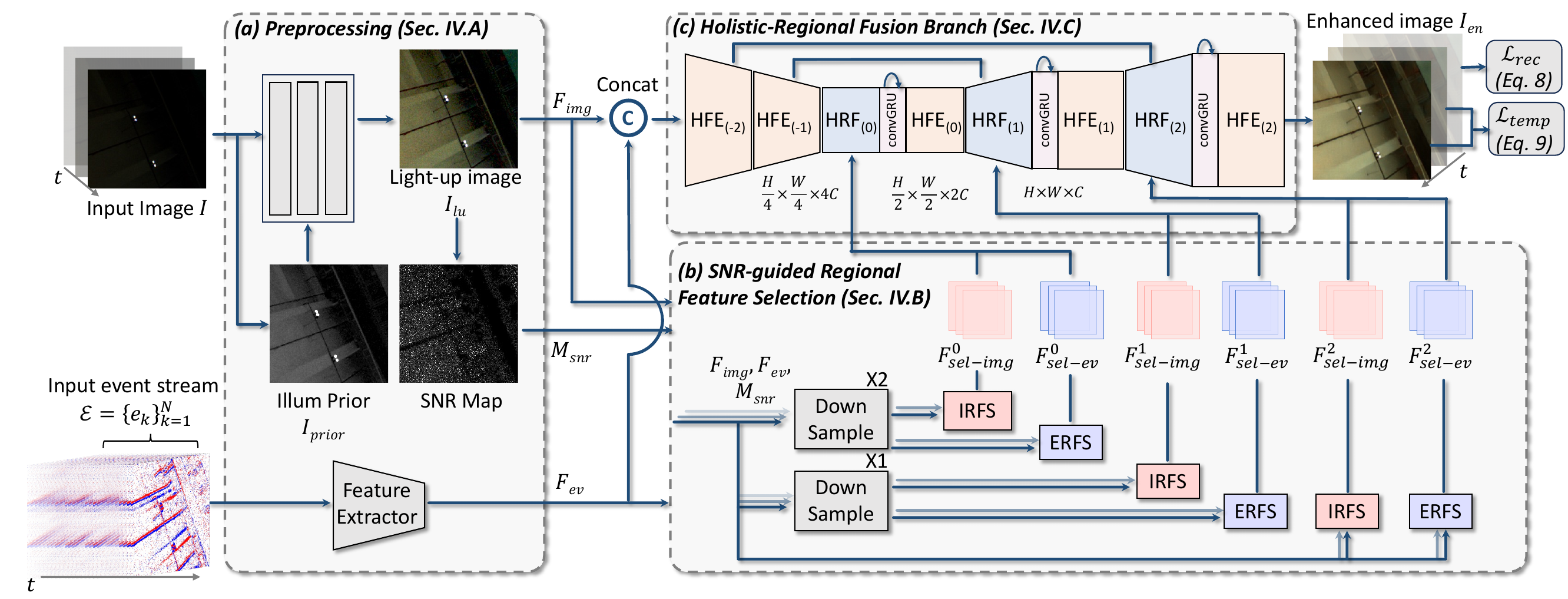}
    \caption{\small \textbf{An overview of our EvLight++ framework}. Our method consists of three parts, \textbf{(a)} Preprocessing (Sec.~\ref{sec:preprocessing}), \textbf{(b)} SNR-guided Regional Feature Selection (Sec.~\ref{sec:regional}), and \textbf{(c)} Holistic-Regional Fusion Branch (Sec.~\ref{sec:holistic}). Specifically, SNR-guided Regional Feature Selection consists of two parts: Image-Regional Feature Selection (IRFS) and Event-Regional Feature Selection (ERFS). Additionally, Holistic-Regional Fusion Branch encompasses Holistic Feature Extraction (HFE), Holistic-Regional Feature Fusion (HRF) and temporal holistic feature enhancement.}
    \vspace{-8pt}
    \label{fig:framework}
\centering
\end{figure*}

\noindent\textbf{Monocular Depth Estimation:} 
For depth annotations, we utilize the Depth Anything V2~\cite{depth_anything_v2} foundation model. DAM is trained on large-scale datasets using a large-scale architecture, making it a powerful foundation model for monocular depth estimation. Following the manual annotation of the camera rotation for each sequence, we generate robust and fine-grained depth maps. Fig.~\ref{fig:data_sample} (b, second row) presents visualization samples of the depth annotations, showcasing the clear structure in various scenes.

\section{The Proposed EvLight++ Framework}
\label{sec:method}

Based on our SDE dataset, we propose a novel event-guided low-light video enhancement framework, called \textbf{EvLight++}, as depicted in Fig.~\ref{fig:framework}.
Our goal is to selectively fuse the features of the image and events to achieve robust performance for event-guided LIE.
EvLight++ takes the low-light sequence $\{\vv{I}_t\}_{t=1}^{T}$ and corresponding event streams $\{\mathcal{E}_t\}_{t=1}^{T}$ with $N$ events as inputs and outputs the enhanced sequence $\{\vv{I}_{en,t}\}_{t=1}^{T}$. \textit{For simplicity, we omit the notation of $t$ in the following explanation}.
Our pipeline mainly consists of three components:
\textbf{1)} Preprocessing,
\textbf{2)} SNR-guided Regional Feature Selection, and
\textbf{3)} Holistic-Regional Fusion Branch.

\noindent \textbf{Event Representation.}
Given an event stream $\mathcal{E}_t=\left\{\vv{e}_k \right\}_{k=1}^N$, we follow \cite{rebecq2019events} to obtain the event voxel grid $\vv{E}$ by assigning the polarity of each event to the two closest voxels. The bin size is set to 32 in all the experiments.

\vspace{-5pt}
\subsection{Preprocessing}
\label{sec:preprocessing}
\noindent \textbf{Initial Light-up.}
As demonstrated in recent frame-based LIE methods~\cite{cai2023retinexformer,wang2023low8,xu2023low7}, coarsely enhancing the low-light image benefits the image restoration process and further boosts the performance.
For simplicity, we follow Retinexformer~\cite{cai2023retinexformer} for the initial enhancement.
As shown in the Fig.~\ref{fig:framework}, we estimate the initial light-up image $\vv{I}_{lu}$ as:
\begin{equation}
\vspace{-5pt}
    \vv{I}_{lu} = \vv{I} \odot \vv{L}, \\ \vv{L} = \mathcal{F}(\vv{I}, \vv{I}_{prior}),
\end{equation}
where $\vv{I}_{prior}=max_c(\vv{I})$ denotes the illumination prior map, with $max_c$ denoting the operation that computes the max values for each pixel across channels.
$\mathcal{F}$ outputs the estimated illumination map $\vv{L}$, which is then applied to the input image $\vv{I}$ through a pixel-wise dot product. 

\noindent \textbf{The SNR Map.}
Following the previous approaches~\cite{buades2005non,dabov2006image,xu2022snr4}, we estimate the SNR map based on the initial light-up image $\vv{I}_{lu}$ and make it an effective prior for the SNR-guided regional feature selection in Sec.~\ref{sec:regional}.
Given the initial light-up image $\vv{I}_{lu}$, we first convert it into grayscale one $\vv{I}_{g}$, \ie, ${\vv{I}}_{g}\in \mathbb{R}^{H\times W}$, followed by computing the SNR map $\vv{M}_{snr} = \tilde{\vv{I}}_g / abs(\vv{I}_{g} - \tilde{\vv{I}}_g)$, where $\tilde{\vv{I}}_g$ is the denoised counterpart of $\vv{I}_{g}$. 
In practice, similar to SNR-Net~\cite{xu2022snr4}, the denoised counterpart is calculated with the mean filter.


\noindent \textbf{Feature Extraction.} 
Image feature $\vv{F}_{img}$ of light-up image $\vv{I}_{lu}$ and event feature $\vv{F}_{ev}$ of the event voxel grid $\vv{E}$ are initially extracted with $conv3\times3$. 

\subsection{SNR-guided Regional Feature Selection}
\label{sec:regional}

In this section, we aim to \textit{selectively extract the regional features from either images or events}.
We design an image-regional feature selection (IRFS) block to select image feature with higher SNR values, thereby obtaining image-regional feature, less affected by noise.
However, SNR map assigns low SNR values to not only high-noise regions but also edge-rich regions.
Consequently, solely extracting features from regions with high SNR values can inadvertently suppress edge-rich regions.
To address this, we introduce an event-regional feature selection (ERFS) block for enhancing edges in areas with poor visibility and high noise.

As shown in Fig.~\ref{fig:framework}, inputs of this module include the image feature $\vv{F}_{img}$, the event feature $\vv{F}_{ev}$, and the SNR map $\vv{M}_{snr}$.
Firstly, the image feature $\vv{F}_{img}$ and event feature $\vv{F}_{ev}$ are down-sampled with $conv4\times4$ layers with the stride of 2 and SNR map $\vv{M}_{snr}$ undergoes an averaging pooling with the kernel size of 2.
These donwsampling operations are represented as `\textit{Down Sample}' in Fig.~\ref{fig:framework} and we attain different scale image feature $\vv{F}^i_{img} \in \mathbb{R}^{\frac{H}{2^{2-i}} \times \frac{W}{2^{2-i}} \times 2^{2-i}C}$, event feature $\vv{F}^i_{ev} \in \mathbb{R}^{\frac{H}{2^{2-i}} \times \frac{W}{2^{2-i}} \times 2^{2-i}C}$, and SNR map $\vv{M}^{i}_{snr} \in \mathbb{R}^{\frac{H}{2^{2-i}} \times \frac{W}{2^{2-i}}}$  where $i = 0,1,2$.
Then, the image feature $\vv{F}^i_{img}$ and event feature $\vv{F}^i_{ev}$ are selected with the guidance of SNR map $\vv{M}^i_{snr}$ in IRFS block, and ERFS block. 
These two blocks then output the selected image features $\vv{F}^{i}_{sel-img}$ and event features $\vv{F}^{i}_{sel-ev}$, respectively. 
We now describe the details of these two blocks.

\begin{figure}[t]
\centering
    \includegraphics[width=1\linewidth]{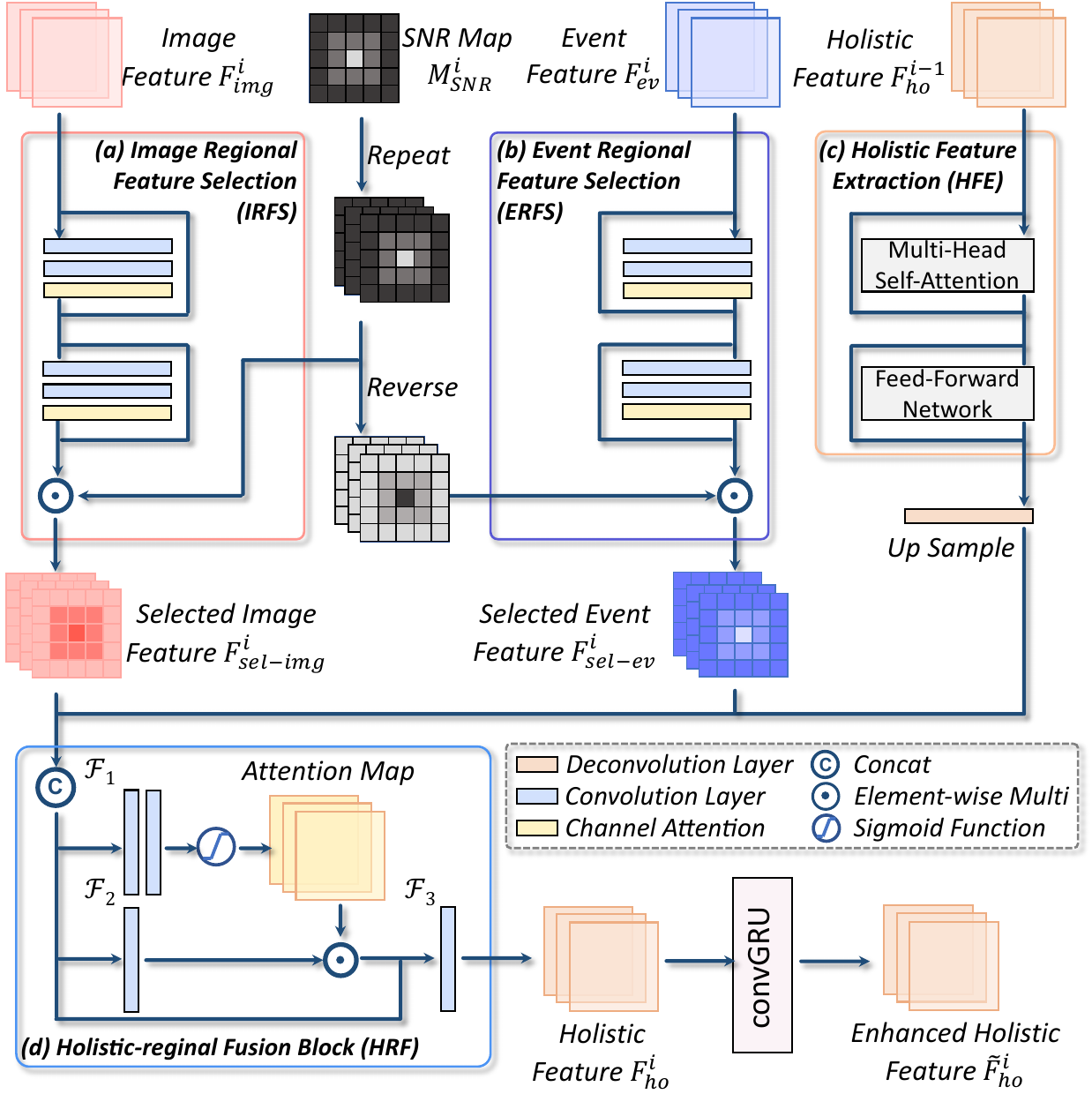}
    \vspace{-8pt}
    \caption{Details of each block in SNR-guided Regional Feature Selection and Holistic-Regional Fusion Branch's decoder.}
    \label{fig:snr_block}
\centering
\end{figure}

\noindent \textbf{Image-Regional Feature Selection (IRFS) Block.} 
As depicted in Fig.~\ref{fig:snr_block} (a), for an image feature $\vv{F}^i_{img}$, we initially process it through two residual blocks~\cite{he2016deep} to extract regional information and yield the output $\hat{\vv{F}}^i_{img}$.
Each block comprises two $conv3\times3$ layers and an efficient channel attention layer~\cite{wang2020eca}. 
The SNR map $\vv{M}^i_{snr}$ is then expanded along the channel to align with the image feature's channel dimensions. Then, we normalize it and make it within the range of $[0, 1]$. 
We then apply a predefined threshold on the SNR map to attain $\hat{\vv{M}}^{i}_{snr}$.
To emphasize regions with higher SNR values and obtain the selected image feature $\vv{F}^{i}_{sel-img}$, we perform an element-wise multiplication $\odot$ between the extended SNR map and the image feature $\hat{\vv{F}}^i_{img}$, as follows:
\begin{equation}
\vspace{-1pt}
    \begin{aligned}
    \vv{F}^{i}_{sel-img} &= \hat{\vv{M}}^{i}_{snr} \odot \hat{\vv{F}}^i_{img}. \\
    \end{aligned}
\end{equation}



\noindent \textbf{Event-Regional Feature Selection (ERFS) Block.}
Edge-rich regions in the initial light-up image, particularly those underexposed, exhibit low SNR values.
Additionally, we observe that events in high SNR regions (\eg well-illuminated smooth planes) predominantly leak noise and shot noise events. 
Consequently, we design the ERFS block that utilizes the inverse of the SNR map to selectively enhance edges in low-visibility, high-noise areas, and to suppress noise events in sufficiently illuminated regions.
The initial processing in this block follows a similar architecture to that used for the IRFS block, with $\vv{F}^i_{ev}$ as the input and $\hat{\vv{F}}^i_{ev}$ as the output.
Given the SNR map $\hat{\vv{M}}^{i}_{snr}$, we obtain the reserve of SNR map $\vv{\bar{M}}^{i}_{snr}$ by  $\mathds{1}$ - $\hat{\vv{M}}^{i}_{snr}$. To obtain the selected event-regional feature $\vv{F}^{i}_{sel-ev}$, the element-wise multiplication product $\odot$ between the reserve of SNR map and the event feature is carried out, which is formulated as:
\begin{equation}
\vspace{-2pt}
    \begin{aligned}
    \vv{F}^{i}_{sel-ev} &= \vv{\bar{M}}^{i}_{snr} \odot \hat{\vv{F}}^i_{ev}. \\
    \end{aligned}
\end{equation}

\subsection{Holistic-Regional Fusion Branch}
\label{sec:holistic}
In this section, we aim to \textit{extract holistic features from both event and image features to establish long-range channel-wise dependencies}.
Besides, the holistic features are enhanced with the selected image-regional and event-regional features in the holistic-region feature fusion process.

Fig.~\ref{fig:framework} (c) depicts our holistic-regional fusion branch, which employs a UNet-like architecture~\cite{ronneberger2015u} with the skip connections.
This branch takes the concatenated feature of image $\vv{F}_{img}$ and event $\vv{F}_{ev}$ from the preprocessing stage (Sec.~\ref{sec:preprocessing}) as the input and the enhanced image $\vv{I}_{en}$ as the output.
In the contracting path, there are 2 layers and the output of each layer is ${\vv{F}}^{i+1}_{ho} \in \mathbb{R}^{\frac{H}{2^{2-|i+1|}} \times \frac{W}{2^{2-|i+1|}} \times 2^{2-|i+1|}C}$ where $i = -2,-1$.
In the $i$-th layer, the holistic feature $\vv{F}^i_{ho}$ first undergoes the holistic feature extraction (HFE) block.
Then with a strided $conv4\times4$ down-sampling operation, the holistic feature $\vv{F}^{i+1}_{ho}$ is obtained.
In the expansive path, the output of each layer is ${\vv{F}}^i_{ho}$ where $i = 0, 1, 2$. 
As shown in Fig.~\ref{fig:snr_block}, the holistic feature ${\vv{F}}^{i-1}_{ho}$ is processed with the HFE block and $\hat{\vv{F}}^{i-1}_{ho}$ is produced.
Then, the holistic feature $\hat{\vv{F}}^{i-1}_{ho}$ is up-sampled with a strided $deconv2\times2$ and it is fused with the selected regional image $\vv{F}^{i}_{sel-img}$ and event features $\vv{F}^{i}_{sel-ev}$ in the holistic-regional fusion (HRF) block.

\noindent \textbf{Holistic Feature Extraction (HFE) Block.} 
As shown in Fig.~\ref{fig:snr_block} (c), holistic feature extraction is mainly composed of a multi-head self-attention module and a feed-forward network.
Given a holistic feature ${\vv{F}}^{i-1}_{ho}$, it is processed as follows:
\begin{equation}
\vspace{-5pt}
    \begin{aligned}
       \hat{\vv{F}}^{i-1}_{mid} &= \text{Attention}({\vv{F}}^{i-1}_{ho})+{\vv{F}}^{i-1}_{ho}, \\
       \hat{\vv{F}}^{i-1}_{ho} &= 
       \text{FFN}(\text{LN}(\hat{\vv{F}}^{i-1}_{mid})) + 
       \hat{\vv{F}}^{i-1}_{mid},
    \end{aligned}
\end{equation}
where $\hat{\vv{F}}^{i-1}_{mid}$ is the middle output, LN is the layer normalization, FFN represents the feed-forward network, and Attention signifies the channel-wise self-attention, analogous to the multi-head attention mechanism employed in~\cite{zamir2022restormer}.

\noindent \textbf{Holistic-Regional Fusion (HRF) Block.} 
This block first concatenates the selected image features $\vv{F}^i_{sel-img}$, selected event features $\vv{F}^i_{sel-ev}$, and up-sampled holistic features $\hat{\vv{F}}^{i-1}_{ho}$.
This concatenated feature $\vv{F}^i_{cat}$ is then passed through $conv3\times3$ layers to generate a spatial attention map.
Sequentially, an element-wise multiplication is applied between the attention map and the concatenated features, as follows:
\begin{equation}
\vspace{-5pt}
    \begin{aligned}
       {\vv{F}}^i_{ho} &= \mathcal{F}_3(\sigma(\mathcal{F}_1(\vv{F}^i_{cat})) \odot \mathcal{F}_2(\vv{F}^i_{cat}) + \vv{F}^i_{cat}),
    \end{aligned}
\end{equation}
where $\mathcal{F}_i$ is the convolution operation indicated in Fig.~\ref{fig:snr_block} (d). $\sigma$ and $\odot$ denote the Sigmoid function and the element-wise production, respectively.

\noindent \textbf{Temporal Holistic Feature Enhancement.}
\label{sec:recurrent} 
Previous video enhancement methods~\cite{wang2021seeing21,liang2023coherent} typically model temporal dependencies using multi-frame fusion approaches. However, this parallel processing approach requires substantial GPU memory and computational resources, especially within our attention-based framework. Additionally, extracting temporal information from asynchronous events in batch mode is limited, given the high temporal resolution of event cameras. To address these challenges, we propose a recursive pipeline that effectively models temporal information while reducing GPU costs.
Specifically, we process the holistic feature $\vv{F}^i_{ho}$ using a convolutional gated recurrent unit (convGRU)~\cite{ballas2015delving}. 
The convGRU block employs an updatable latent state $\vv{S}$ to capture temporal information. By integrating the holistic feature $\vv{F}^i_{ho}$ with the previous state $\vv{S}_{t-1}$, it generates both the temporal feature $\vv{\tilde{F}}^i_{ho}$ and the current state $\vv{S}_{t}$, as described by the following equation:
\begin{equation}
    \vv{\tilde{F}}^i_{ho}, \vv{S}_{t} = convGRU(\vv{F}^i_{ho}, \vv{S}_{t-1})
\end{equation}
where $\vv{S}_{t}$ is the temporal information for the $t$-th prediction.

\subsection{Optimization}
\label{sec:loss}
The loss function $\mathcal{L}$ used for training our EvLight++ model is composed of a reconstruction loss and a temporal loss, formulated as:
\begin{equation}
    \mathcal{L} = \mathcal{L}_{rec} + \lambda_{temp} \mathcal{L}_{temp},
\end{equation}
where $\lambda_{temp}=1$ is a hyper-parameter that balances the contribution of the temporal term. The reconstruction term is designed to ensure fidelity between the enhanced frame $\vv{I}_{en}$ and the GT $\vv{I}_{gt}$ at both the pixel and feature levels:
\begin{equation}
    \mathcal{L}_{rec} = \sqrt{||\vv{I}_{en}-\vv{I}_{gt}||^2 +\epsilon^2} + \lambda||\Phi({\vv{I}}_{en}) - \Phi({\vv{I}}_{gt}) ||_1 , 
\end{equation}
where $\lambda=0.5$ is a hyper-parameter, $\epsilon$ is set to $10^{-4}$, and $\Phi$ represents feature extraction using the AlexNet~\cite{krizhevsky2012imagenet}.

The temporal term is included to regularize temporal fidelity, enhancing the consistency of illumination or appearance over time:
\begin{equation}
    \mathcal{L}_{temp} = ||(\vv{I}_{en,t} - \vv{I}_{en,t-1}) - (\vv{I}_{gt,t} - \vv{I}_{gt,t-1})||_1,
\end{equation}
where $\vv{I}_{*,t}$ denotes the $t$-th frame in the sequence.

\section{Experiments}
\subsection{Experimental Settings}
\noindent\textbf{Implementation Details:}
We employ the Adam optimizer~\cite{kingma2014adam} for all experiments, with learning rates of $1e-4$ and $2e-4$ for SDE and SDSD datasets, respectively. 
Our framework is trained for 80 epochs with a batch size of 8 using an NVIDIA A30 GPU. 
We apply random cropping, horizontal flipping, and rotation for data augmentation. 
The cropping size is 256 $\times$ 256, and the rotation angles include 90, 180, and 270 degrees.

\begin{figure}[!]
\centering
    \includegraphics[width=0.95\linewidth]{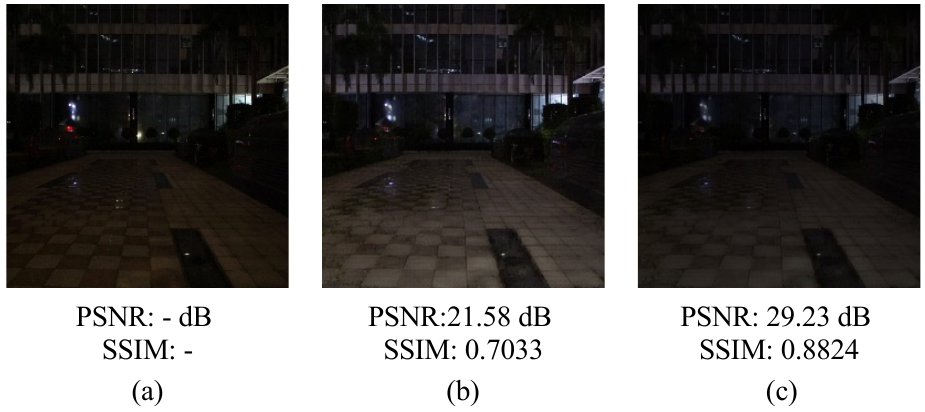}
    \caption{Illustration of the impact of overall brightness fine-tuning on the calculation of PSNR and SSIM. (a) displays the GT image, (b) shows the initial enhanced result, and (c) presents the enhanced result after applying overall brightness fine-tuning.
}
    \label{fig:psnr_star}
\centering
\end{figure}

\noindent\textbf{Evaluation Metrics:} We use the peak-signal-to-noise ratio (PSNR)~\cite{hore2010image} and SSIM~\cite{wang2004image} for evaluation.
In the calculation of PSNR and SSIM, the distance of light level between the predicted image and GT image is also accounted for, while there is no ground-truth light level for image enhancement partice~\cite{wu2022uretinex5}.
We notice that the predicted image with better visual experience (\eg brighter) than the ground-truth image will be assigned low PSNR and SSIM due to the distinctions in light level as shown in Fig.~\ref{fig:psnr_star}.
Following the finetuning of the overall brightness of predicted results in previous methods~\cite{zhang2019kindling13,wu2023learning}, we introduce the PSNR* as the metric to assess image restoration effectiveness beyond light fitting.
The calculation of PSNR* is formulated as:
\begin{equation}
    \begin{aligned}
    \vspace{-5pt}
    \text{PSNR*} &= \text{PSNR}(\vv{I}_{en}\times{\vv{R}}_{gt-en},\vv{I}_{gt}), \\
    {\vv{R}}_{gt-en} &= \text{Mean}(\text{Gray}(\vv{I}_{gt})) / \text{Mean}(\text{Gray}(\vv{I}_{en})),
    \end{aligned}
\end{equation}
where $\vv{I}_{en}$, $\vv{I}_{gt}$, Gray, Mean, and PSNR represent the enhanced image, the ground-truth image, the operation of converting RGB images to grayscale ones, the operation of getting mean value, and the operation of calculating PSNR value, respectively. 

\begin{table*}[t]
    \centering
    \caption{\textbf{Comparisons on our SDE dataset and SDSD~\cite{wang2021seeing21} dataset}. The highest result is highlighted in \textbf{bold} while the second highest result is highlighted in \underline{underline}. Since E2VID+~\cite{reducingsimtoreal} can only reconstruct grayscale images, its metrics are calculated in grayscale.}
    \setlength{\tabcolsep}{3pt}
    \resizebox{1.0\textwidth}{!}{
\begin{tabular}{cccccccccccccc}
\hline
\multirow{2}{*}{Input} & \multirow{2}{*}{Method} & \multicolumn{3}{c}{SDE-in}  & \multicolumn{3}{c}{SDE-out}   & \multicolumn{3}{c}{SDSD-in}    & \multicolumn{3}{c}{SDSD-out}\\
\cmidrule(r){3-5} \cmidrule(r){6-8} \cmidrule(r){9-11} \cmidrule(r){12-14}
& & \multicolumn{1}{l}{PSNR$\uparrow$}  & \multicolumn{1}{l}{PSNR*$\uparrow$} & \multicolumn{1}{l}{SSIM$\uparrow$}  & \multicolumn{1}{l}{PSNR$\uparrow$} & \multicolumn{1}{l}{PSNR*$\uparrow$} & \multicolumn{1}{l}{SSIM$\uparrow$}  & \multicolumn{1}{l}{PSNR$\uparrow$} & \multicolumn{1}{l}{PSNR*$\uparrow$} & \multicolumn{1}{l}{SSIM$\uparrow$}  & \multicolumn{1}{l}{PSNR$\uparrow$} & \multicolumn{1}{l}{PSNR*$\uparrow$} & \multicolumn{1}{l}{SSIM$\uparrow$}  \\ \hline
Event Only                   & E2VID+ (ECCV'20) \cite{reducingsimtoreal}           
&15.19              &15.92              &0.5891                                &15.01              &16.02                    &0.5765                            &13.48                    &13.67              &0.6494                      
&16.58      &17.27       &0.6036     \\ \hline
\multirow{4}{*}{Image Only}  & SNR-Net (CVPR'22) \cite{xu2022snr4}          &20.05         &21.89        &0.6302                       &22.18         &22.93 &0.6611                      &24.74         &25.30 &0.8301                            &24.82         &26.44 &0.7401                 \\
                            & Uformer (CVPR'22) \cite{wang2022uformer}          &21.09         &22.75       &{0.7524}                      &22.32         &23.57 &{0.7469}                      &24.03         &25.59 &{0.8999}                            &24.08         &25.89 &{0.8184}                 \\
                             & LLFlow-L-SKF (CVPR'23)~\cite{wu2023learning}     &20.92         &22.22 &0.6610                        &21.68         &23.41 &0.6467                       &23.39              &24.13 &0.8180                              &20.39         &24.73 &0.6338               \\
                             & Retinexformer (ICCV'23)~\cite{cai2023retinexformer}    &21.30        &23.78  &0.6920                       &{22.92}         &23.71 &0.6834                       &25.90         &25.97 &0.8515                             &{26.08}         &{28.48} &0.8150                 \\ \hline
\multirow{3}{*}{Image+Event} & ELIE (TMM'23)~\cite{jiang2023event}              &19.98         &21.44 &0.6168                          &20.69         &23.12 &0.6533                      &{27.46}         &{28.30} &{0.8793}                             &23.29              &28.26 &0.7423          \\
                             & eSL-Net (ECCV'20)~\cite{wang2020event}         &21.25       &23.19 &0.7277                        &22.42        &{24.39} &0.7187                              &24.99         &25.72 &0.8786                         &24.49          &26.36 &0.8031             \\
                             & Liu \etal (AAAI'23)~\cite{liu2023low}         &{21.79}       &{23.88} &0.7051                            &22.35        &23.89 &0.6895                              &{27.58}         &{28.43} &{0.8879}                              &23.51          &27.63 &0.7263                 \\
                             & EvLight (CVPR'24)~\cite{liang2024towards}                       &\underline{22.44}          &\underline{24.81} &\underline{0.7697}                           &\underline{23.21}        &\underline{25.60} &\underline{0.7505}                     &\underline{28.52}       &\underline{29.73} &\underline{0.9125}                        &\underline{26.67}        &\underline{30.30} &\underline{0.8356}
                             \\ \hline

                             \multirow{2}{*}{Video+Event} &
                             EvLowlight(ICCV'23)~\cite{liang2023coherent} &20.57 & 22.14 & 0.6217 & 22.04&23.72&0.6485&23.14&24.95&0.8143&23.27&28.11&0.7363\\
                             &EvLight++ (Ours)&\textbf{22.67}&\textbf{25.83}&\textbf{0.7791}& \textbf{23.34}&\textbf{26.01}&\textbf{0.7676}&\textbf{29.61}&\textbf{30.52}&\textbf{0.9184}&\textbf{27.52}&\textbf{30.59}&\textbf{0.8635}\\
                             \hline
\multicolumn{1}{l}{}         & \multicolumn{1}{l}{}    & \multicolumn{1}{l}{}     & \multicolumn{1}{l}{}     & \multicolumn{1}{l}{}      & \multicolumn{1}{l}{}     & \multicolumn{1}{l}{}     & \multicolumn{1}{l}{}      & \multicolumn{1}{l}{}     & \multicolumn{1}{l}{}                          
\end{tabular}
    }
    
    \label{tab:main_result}
\end{table*}



\begin{figure*}[!]
\centering
    \includegraphics[width=1\linewidth]{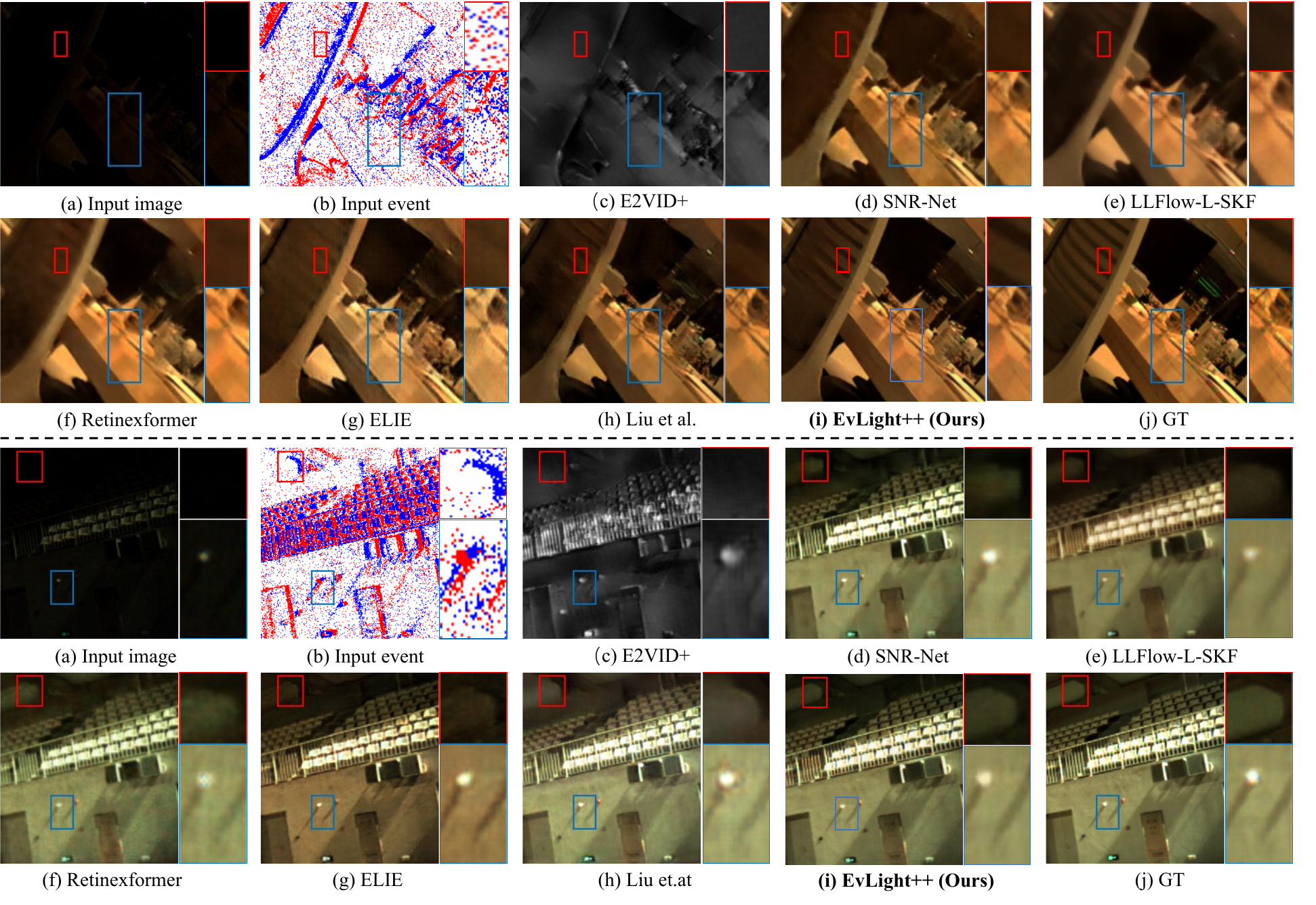}
    \caption{Qualitative results on our SDE-in and SDE-out dataset. For each dataset, we provide the results of a typical scene for selected compared methods. (Best viewed on screen with zoom).}
    \vspace{-8pt}
    \label{fig:visual-sde}
\centering
\end{figure*}

\subsection{Datasets}
\noindent\textbf{1) SED dataset}
contains 91 image+event paired sequences captured with a DAVIS346 event camera~\cite{scheerlinck2019ced} which outputs RGB images and events with the resolution of $346\times260$.
For all collected sequences, 76 sequences are selected for training, and 15 sequences are for testing. 

\noindent\textbf{2) SDSD dataset}~\cite{wang2021seeing21}~
provides paired low-light/normal-light videos with $1920\times1080$ resolution containing static and dynamic versions.
We choose the dynamic version for simulating events and employ the same dataset split scheme as in SDSD: 125 paired sequences for training and 25 paired sequences for testing.
We first downsample the original videos to the same resolution ($346\times260$) of the DAVIS346 event camera. 
Then, we input the resized images to the event simulator v2e~\cite{hu2021v2e} to synthesize event streams with noise under the default noisy model.

\begin{figure*}[!]
\centering
    \includegraphics[width=1\linewidth]{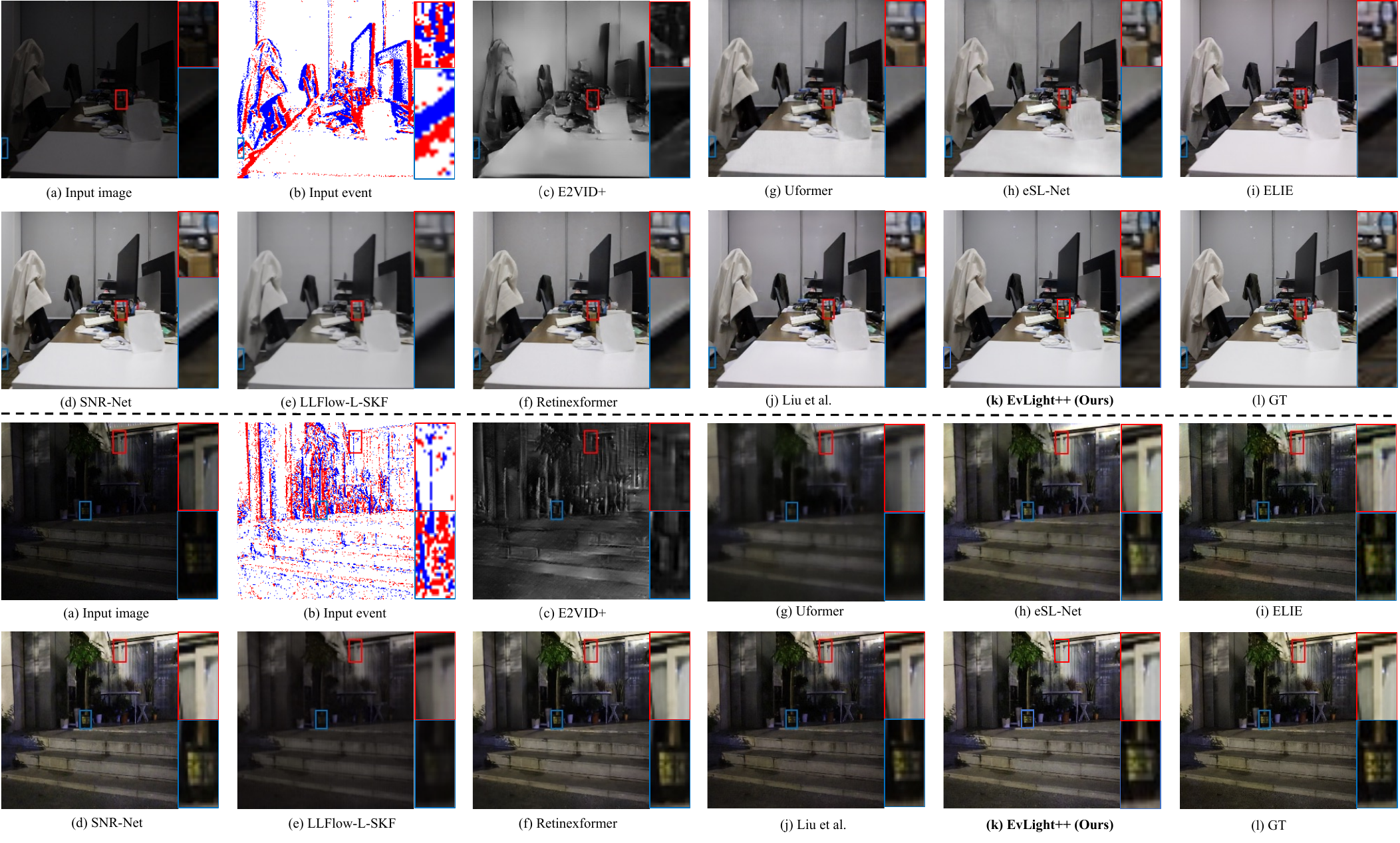}
    \caption{Qualitative results on SDSD-in and SDSD-out dataset~\cite{wang2021seeing21}. We provide the results of a typical scene for selected compared methods. (Best viewed on screen with zoom).}
    \vspace{-10pt}
    \label{fig:visual-sdsd}
\centering
\end{figure*}

\subsection{Comparison and Evaluation}
We compare our method with recent methods with four different settings: 
\textbf{(I)} the experiment with events as input, including E2VID+~\cite{reducingsimtoreal}. \textbf{(II)} the experiment with a RGB image as input, including SNR-Net~\cite{xu2022snr4}, Uformer~\cite{wang2022uformer}, LLFlow-L-SKF~\cite{wu2023learning}, and Retinexformer~\cite{cai2023retinexformer}.
\textbf{(III)} the experiment with a RGB image and paired events as inputs, including ELIE~\cite{jiang2023event}, eSL-Net~\cite{wang2020event}, and Liu \etal~\cite{liu2023low}.
We reproduced ELIE~\cite{jiang2023event} and Liu \etal~\cite{liu2023low} according to the descriptions in the papers, while the others are retrained with the released code.
We replace the event synthesis module in~\cite{liu2023low} by inputting events captured with the event camera or generated from the event simulator~\cite{hu2021v2e}.
\textbf{(IV)} the experiment with video and events as inputs, including EvLowLight~\cite{liang2023coherent}. Since some of the comparison methods don't release their training code, we reproduce the training pipeline following the hyper-parameters as illustrated in the reports.

\subsubsection{Comparison on our SDE Dataset:}
Quantitative results in Tab.~\ref{tab:main_result} demonstrate EvLight++'s superior performance on the SDE dataset, outperforming EvLight with a higher PSNR* by 1.02 dB for SDE-in and 0.41 dB for SDE-out, highlighting the effectiveness of the temporal information. Compared to other baseline methods, our method also delivers better results across all error metrics, with higher PSNR by 0.88 dB for SDE-in and 0.42 dB for SDE-out, demonstrating the superiority of our proposed SNR-guided framework.

To better illustrate the performance of our method, several representative baseline methods are compared qualitatively in Fig.~\ref{fig:visual-sde} for indoor (first two rows) and outdoor (last two rows) scenes, respectively.
For the indoor sample, our method effectively reconstructs clear edges in dark areas (\eg the red box areas), outperforming frame-based methods like Retinexformer~\cite{cai2023retinexformer} and event-guided approaches like Liu~\etal~\cite{liu2023low}. Additionally, in challenging outdoor scenes (\eg the wall in Fig.~\ref{fig:visual-sde}, last two rows), our method exhibits less color distortion and noise than LLFlow-L-SKF~\cite{wu2023learning}, ELIE~\cite{jiang2023event}, and Retinexformer, highlighting the robustness of our approach.

\subsubsection{Comparison on the SDSD Dataset}
Following previous image-only baseline methods, we also conducted comparisons on the SDSD dataset~\cite{wang2021seeing21}. Quantitative outcomes, detailed in Tab.~\ref{tab:main_result}, evaluate our method's effectiveness and generalization. Our method significantly outperforms baselines in all error metrics, leading by more than 1.09 dB for SDSD-in and 0.85 dB for SDSD-out in PSNR.
Our method outperforms baselines significantly in all the error metrics, leading by more than {1.09 dB} for SDSD-in and {0.85 dB} for SDSD-out on PSNR. 
Although ELIE and Liu~\etal~\cite{liu2023low} surpass frame-based methods on the SDSD-in dataset, they suffer from overfitting on the SDSD-out dataset, as evidenced by the substantial disparity between PSNR and PSNR*.
Qualitatively, as shown in Fig.~\ref{fig:visual-sdsd}, our method effectively restores underexposed images to reveal more detailed structures, as highlighted in the red box and blur box areas. Notably, ELIE~\cite{jiang2023event} tends to produce color distortions, as visible in the blue box area of Fig.~\ref{fig:visual-sdsd} (d).

\subsubsection{Performance with higher resolution.}
\begin{figure}[!]
\centering
    \includegraphics[width=1\linewidth]{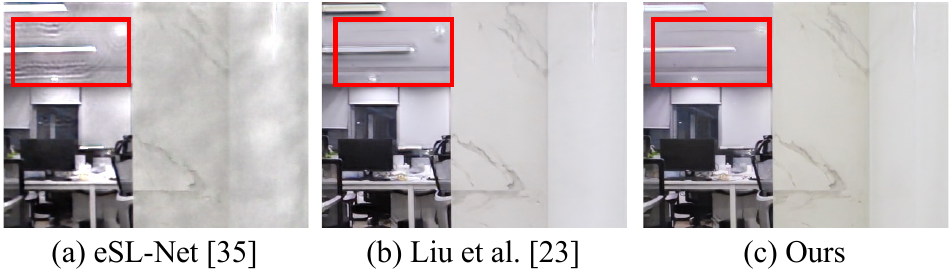}
    \caption{Visual results on SDSD~\cite{wang2021seeing21} with the resolution of $512\times512$.}
    \label{fig:hdresults}
\centering
\end{figure}

Since DAVIS346 event camera only captures low-resolution sequences of $346\times260$, it may restrict the capability of the models. Therefore, we further explore our performance with the higher resolution of $512\times512$, by following the same process in the main paper to synthesize data with SDSD dataset~\cite{wang2021seeing21}. 
As shown in Tab.~\ref{tab:hdresult} and Fig.~\ref{fig:hdresults}, our method achieves the {SOTA} performance and the {best} visual result (see {red box} in Fig.~\ref{fig:hdresults}). 

\begin{table}[!]
\centering
\caption{Numerical results on SDSD~\cite{wang2021seeing21} with a resolution of $512\times512$.}
\resizebox{0.48\textwidth}{!}{
\begin{tabular}{cccc}
\hline
Method    & eSL-Net~\cite{wang2020event}      & Liu \etal~\cite{liu2023low}         & Ours         \\ \hline
PSNR / SSIM & 25.83 / 0.8441 & 28.12 / 0.8791 & \textbf{28.71} / \textbf{0.8923} \\ \hline
\end{tabular}
}
\label{tab:hdresult}
\end{table}

\subsubsection{Generalization Comparison}

\begin{figure}[!]
\centering
    \includegraphics[width=1\linewidth]{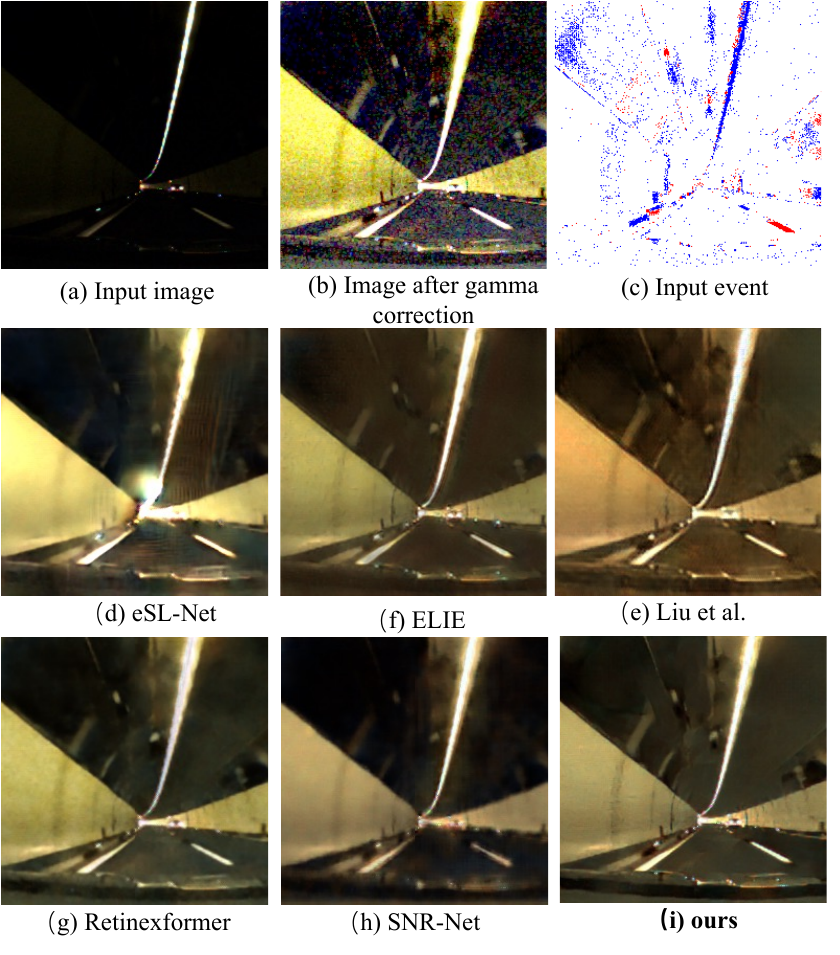}
    \caption{Qualitative results on CED~\cite{scheerlinck2019ced} dataset.
    }
    \label{fig:generlizationced}
\centering
\end{figure}

\begin{figure}[!]
\centering
    \includegraphics[width=1\linewidth]{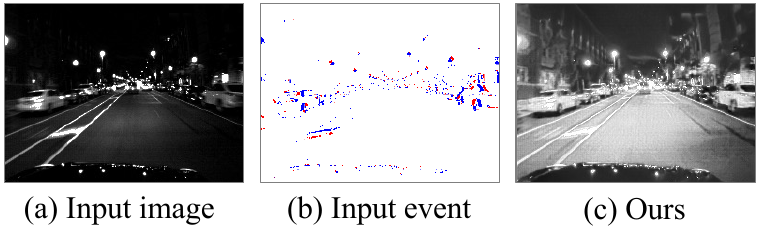}
    \caption{Qualitative results on MVSEC~\cite{zhu2018multivehicle} dataset.
    }
    \label{fig:generlizationmvsec}
\centering
\end{figure}

To assess the generalization capability of our proposed approach, we conducted an experiment using the CED dataset~\cite{scheerlinck2019ced} with the model trained on our SDE dataset. Since the CED dataset lacks paired normal-light images to serve as ground truth, we present qualitative results derived from challenging examples captured from a moving vehicle within a tunnel, under varying lighting conditions, including with and without sunlight.
As depicted in Fig.~\ref{fig:generlizationced}, our method effectively recovers details obscured in low-light images, while simultaneously preventing overexposure, as seen in (d), and color distortion, as seen in (f) and (h).
Additionally, we conducted experiments on the MVSEC dataset~\cite{zhu2018multivehicle}. As shown in Fig.~\ref{fig:generlizationmvsec}, the restored results display clear structures of roads and buildings, and detailed appearances of cars on the road, demonstrating the effectiveness and robustness of our dataset and method.

\subsection{Ablation Studies and Analysis}
We conduct ablation studies on the SDE-in dataset to assess the effectiveness of each module of our method.
The basic implementation, without SNR-guided regional feature selection as described in Sec.~\ref{sec:regional}, is called the \textit{Base} model.

\begin{figure}[!]
\vspace{-10pt}
\centering
    \includegraphics[width=1\linewidth]{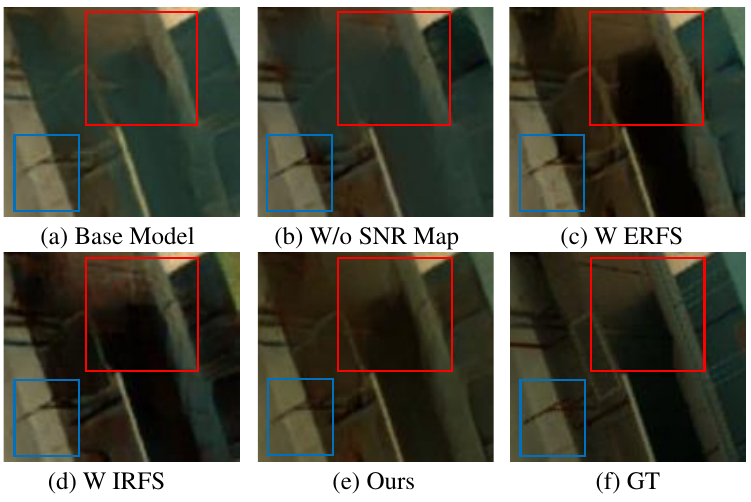}
    \caption{Visualization of ablation results.}
    \label{fig:ablation}
\centering
\end{figure}

\subsubsection{Impact of Events}
To reveal the impact of events, we conduct experiments on the \textit{Base} model. 
The variant excluding events attains a PSNR of 21.35 dB and an SSIM of 0.6985, whereas adding events results in a 0.23 dB improvement in PSNR and a 0.002 improvement in SSIM.
However, the \textit{Base} model cannot fully explore the potential of events demonstrated by the limited improvement in SSIM.

\begin{figure}
    \centering
    \includegraphics[width=1\linewidth]{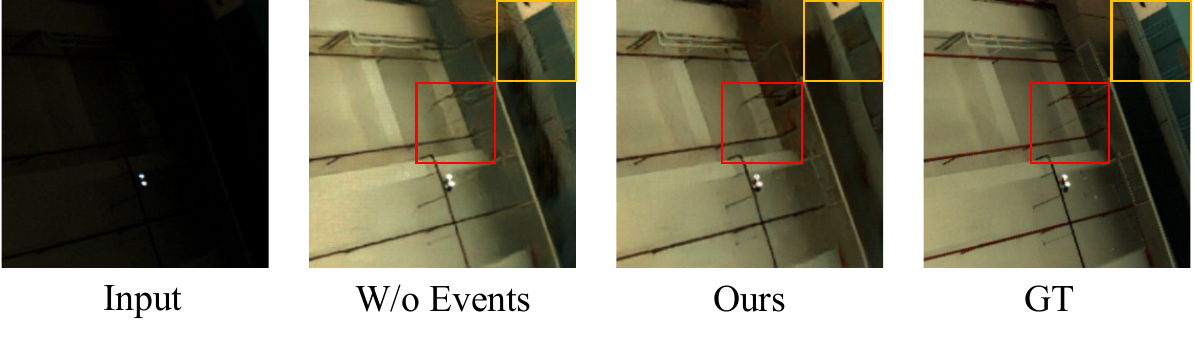}
    \caption{Qualitative results of ablation study on event data.}
    \label{fig:enter-label}
\end{figure}

\subsubsection{Impact of SNR-guided regional feature selection}
We conduct an ablation study in Tab.~\ref{tab:ablation_regioanl}.
Compared with the \textit{Base} model without the whole selection module (\engordnumber{1} row), regional feature selection with an all-ones matrix (\engordnumber{2} row) and SNR-guided regional feature selection (\engordnumber{3} row) yield 0.28 dB and 0.86 dB increase in PSNR, respectively, demonstrating the necessity of regional features and the SNR map.
While both the all-ones matrix and \textit{Base} model show color distortion (e.g., red box in Fig.~\ref{fig:ablation} (a), (b)), (b) provides better structural details than (a).

\begin{table}[!]
\centering
\caption{Ablation of SNR-guided regional feature selection and temporal enhancement.}

\setlength{\tabcolsep}{3pt}
\resizebox{1\linewidth}{!}{
\begin{tabular}{c|ccccc}
\hline
                      & \makecell[c]{Feature Selection} & SNR-guided  & Temporal Enhancement      & PSNR        & SSIM\\ 
\hline 
1             &\ding{55}  &\ding{55}    &\ding{55}   &21.58  &0.7001\\
2             &\ding{51}  &\ding{55}   &\ding{55}     &21.86  &0.7490\\
3             &\ding{51}  &\ding{51} &\ding{55}  &{22.44} &{0.7697}\\
4  &\ding{51}  &\ding{51} &\ding{51} &  \textbf{22.67} & \textbf{0.7791} \\
\hline
\end{tabular}
}
\label{tab:ablation_regioanl}
\end{table}

\subsubsection{Impact of IRFS and ERFS}
To verify them, we conduct an ablation study in Tab.~\ref{tab:ablation_snr}.
Since these modules focus on regional selection within a single frame, the study was performed without temporal enhancement to isolate their impact and demonstrate their effectiveness in region selection.
Compared with the \textit{Base} model (\engordnumber{1} row), image-regional feature selection (IRFS, \engordnumber{2} row), event-regional feature selection (ERFS, \engordnumber{3} row), and the combination of them (\engordnumber{4} row) yields the 0.34 dB, 0.60 dB, and 0.86 dB increase in PSNR, respectively, demonstrating the necessity of the IRFS and ERFS block.
As shown in Fig.~\ref{fig:ablation}, IRFS (d) or ERFS (c) can reduce the color distortion that appears in the \textit{Base} model (a).
With both IRFS and ERFS blocks, our results deliver the best visual quality (\eg red box and blue box in Fig.~\ref{fig:ablation}).

\begin{table}[!]
\centering
\caption{Impact of each module of SNR-guided regional feature selection.}
\setlength{\tabcolsep}{18pt}
\resizebox{1\linewidth}{!}{
\begin{tabular}{c|cccc}
\hline
 &\makecell[c]{IRFS} & \makecell[c]{ERFS} & PSNR & SSIM\\ 
\hline
1 &\ding{55}             &\ding{55}         &21.58  &0.7001 \\
2 &\ding{51}             &\ding{55}         &21.92  &0.7108\\
3 &\ding{55}             &\ding{51}         &22.18  &0.7525\\
4 &\ding{51}             &\ding{51}          &\textbf{22.44}  &\textbf{0.7697}\\
\hline
\end{tabular}
}

\label{tab:ablation_snr}
\end{table}

\subsubsection{Impact of Temporal Enhancement and Loss Function}
Tab.~\ref{tab:ablation_regioanl} (4$^{th}$ row) demonstrates the effectiveness of temporal enhancement. We also conducted an ablation study to evaluate the impact of each component, namely convGRU and temporal loss, as shown in Tab.~\ref{tab:ablation_temporal}. Compared to the original single-frame-based method (EvLight), the inclusion of temporal enhancement using convGRU led to a 0.84 dB improvement in PSNR*. When combined with temporal loss, our EvLight++ achieved further improvements across all evaluation metrics.

\begin{table}[!]
\centering
\caption{Ablation study on temporal enhancement and loss function.}

\setlength{\tabcolsep}{9pt}
\resizebox{1\linewidth}{!}{
\begin{tabular}{c|ccccc}
\hline
                      & \makecell[c]{ConvGRU} & Temporal Loss        & PSNR  &PSNR*      & SSIM\\ 
\hline 
1             &\ding{55}  &\ding{55}       &22.44 & 24.81&0.7697\\
2             &\ding{51}  &\ding{55}        &22.53&25.65&0.7684\\
3             &\ding{51}  &\ding{51}   &\textbf{22.67} & \textbf{25.83}&\textbf{0.7791}\\
\hline
\end{tabular}
}
\label{tab:ablation_temporal}
\end{table}

\subsection{Computational Cost.}
In order to evaluate the computational complexity of our proposed method in contrast to several state-of-the-art (SOTA) methodologies, we present both the floating point operations (FLOPs) and the total number of parameters (Params) in Tab.~\ref{tab:complexity}. 
The computation of FLOPs has been carried out at a resolution of 256 $\times$ 256. 
It is important to note that the channel numbers in Liu \etal~\cite{liu2023low} have not been explicitly stated; therefore, for the purpose of maintaining a fair comparison, we have assumed the channel number in each layer to be identical to that in our method.
In Tab.~\ref{tab:inferencetime}, we compare the inference time of our method with other event-guided LIE methods, all tested on an NVIDIA A30 GPU with mixed precision~\cite{micikevicius2017mixed}.
Our inference time is comparable to that of Liu \etal~\cite{liu2023low} and shorter than ELIE's~\cite{jiang2023event} and eSL-Net's~\cite{wang2020event}.
\begin{table}[!]
\centering
\caption{Comparison of computational complexity.}

\resizebox{1\linewidth}{!}{
\begin{tabular}{llll}
\hline
Input &Method                                         & FLOPs (G) & Params (M) \\ \hline
Event Only                   & E2VID+ (ECCV'20)~\cite{reducingsimtoreal}                         &27.99      &10.71          \\ \hline
\multirow{4}{*}{Image Only}  & SNR-Net (CVPR'22)~\cite{xu2022snr4}                        &26.35    & 4.01           \\
                             & Uformer (CVPR'22)~\cite{wang2022uformer}                        &12.00    & 5.29           \\
                             & LLFlow-L-SKF (CVPR'23)~\cite{wu2023learning}                   &409.50     & 39.91          \\
                             & Retinexformer (ICCV'23)~\cite{cai2023retinexformer}                  &15.57    & 1.61           \\ \hline
\multirow{4}{*}{Image+Event} & ELIE (TMM'23)~\cite{jiang2023event}                            &440.32   & 33.36         \\
                             & eSL-Net (ECCV'20)~\cite{wang2020event}                         &560.94   & 0.56           \\
                             & Liu \etal (AAAI'23)~\cite{liu2023low}                      &44.71    & 47.06          \\
                             & EvLight (CVPR2024)~\cite{liang2024towards}                                   &180.90   & 22.73          \\ \hline
\multirow{1}{*}{Video+Event} & \textbf{EvLight++ (Ours)} & 225.91 & 26.21 \\
\hline
\end{tabular}
}
\label{tab:complexity}
\end{table}

\begin{table}[!]
\centering
\caption{Inference time with the resolution of $256\times256$.}
\setlength{\tabcolsep}{2pt}
\resizebox{1\linewidth}{!}{
\begin{tabular}{ccccc}
\hline
Method         & ELIE~\cite{jiang2023event}      & eSL-Net~\cite{wang2020event} & Liu \etal~\cite{liu2023low}  & \textbf{EvLight++ (Ours)}     \\ \hline
Inference time & 108.42 ms & 37.05 ms & 27.39 ms & 33.54 ms \\ \hline
\end{tabular}
}
\label{tab:inferencetime}
\end{table}

\subsection{Failure Cases Analysis}
\begin{figure}[!]
\centering
    \includegraphics[width=0.95\linewidth]{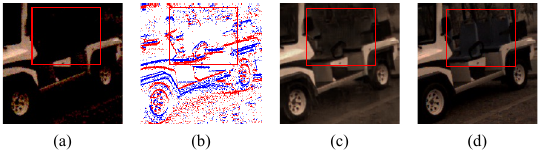}
    \caption{An example of the failure case. (a) shows the low-light image after gamma correction, (b) displays the input low-light events, (c) presents our results, and (d) depicts the GT image.}
    \label{fig:failure_case}
\centering
\end{figure}

As illustrated by the red box in Fig.~\ref{fig:failure_case}, our method does not adequately capture the details of the chairs and the background behind the car. In this scenario, the task of low-light enhancement becomes a highly challenging inpainting problem due to the insufficient color and structural information provided by the input low-light images and events.

\begin{figure*}[!]
    \centering
    \includegraphics[width=1\linewidth]{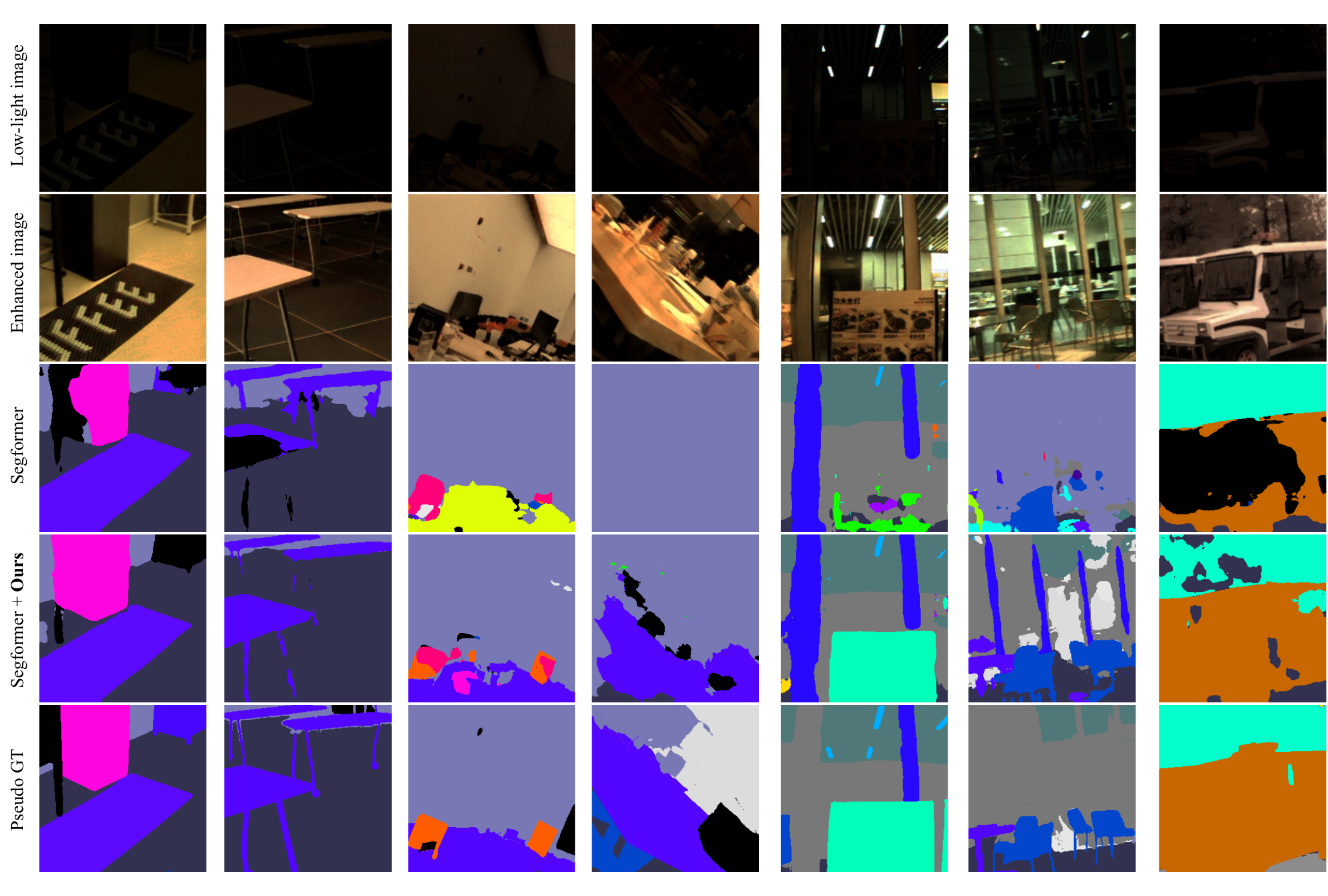}
    \caption{Qualitative comparison of semantic segmentation result for low-light scenes. The baseline method (\ie Segformer~\cite{xie2021segformer}) use the low-light image as input, while ours (\ie Segformer~\cite{xie2021segformer}+Ours) uses the enhanced image as input.}
    \label{fig:downstream_seg}
\end{figure*}

\subsection{Downstream Applications}
\label{sec:downstream_result}
Several methods~\cite{kong2024openess, gehrig2021combining, cannici2019asynchronous} have been proposed to directly apply downstream pipelines to raw event data. However, when frame data is available, it is more robust and flexible to enhance low-quality frames with event data and then feed the high-quality frames to off-the-shelf frame-based algorithms for downstream tasks. This approach allows captured event streams to directly utilize existing, well-established downstream vision algorithms. To further leverage the unique capabilities of event cameras, we conduct extensive experiments on downstream vision tasks.

To ensure a comprehensive comparison, we have provided abundant and precise labels for downstream tasks based on our large-scale paired dataset and well-designed annotation pipeline. To the best of our knowledge, this is the first dataset featuring paired low-light/normal-light videos with simultaneous labels for downstream tasks. To evaluate the effectiveness of the proposed dataset and method, we perform successive vision tasks on two types of images: low-light intensity images and restored high-level images from our EvLight++ trained on the SDE dataset, using the estimated output to calculate error metrics with the labels.

\subsubsection{Semantic Segmentation}
Semantic segmentation is a fundamental task in computer vision that involves segmenting and classifying each pixel in an image. In this task, we apply our method to demonstrate its potential in facilitating the accurate identification of semantic classes, particularly in low-light conditions. To evaluate the performance of our LLE method, we employ the off-the-shelf Segformer~\cite{xie2021segformer} pre-trained on the ADE-20K~\cite{zhou2019semantic} dataset.

For quantitative comparison, we report average accuracy (aAcc), mean Intersection over Union (mIoU), and mean accuracy (mAcc) in Tab.~\ref{tab:downstream_seg}, providing insights into the model's capability to segment semantic masks. With the enhancement provided by our method, most error metrics in semantic segmentation show significant improvement, especially for indoor scenes. The visual results in Fig.~\ref{fig:downstream_seg} demonstrate that our method achieves greater overlap with the ground truth semantic masks, whereas the original Segformer struggles in regions affected by low-light conditions.

\begin{table}[!]
    \centering
    \caption{Quantitative results of semantic segmentation on our SDE dataset. The annotations on the aligned normal-light videos are used as ground truth for calculating evaluation metrics.}
     \setlength{\tabcolsep}{3pt}
    \resizebox{0.5\textwidth}{!}{
    \begin{tabular}{lcccccc}
    \hline
    \multirow{2}{*}{Method} & \multicolumn{3}{c}{SDE-in}& \multicolumn{3}{c}{SDE-out} \\
     \cmidrule(r){2-4} \cmidrule(r){5-7}
         & aAcc$\uparrow$ & mIoU$\uparrow$ & mAcc$\uparrow$ & aAcc$\uparrow$ & mIoU$\uparrow$ & mAcc$\uparrow$\\
       \hline
        SegFormer~\cite{xie2021segformer}&52.51 & 11.02 & 13.55 &  53.74 & 21.58 & 31.02 \\
        SegFormer~\cite{xie2021segformer}+\textbf{Ours} &59.08 & 13.78 &17.47&  56.11 & 23.14 & 30.70 \\
        \hline
        \textit{Improve (\%)} &\textcolor{blue}{+12.51}&\textcolor{blue}{+15.97}&\textcolor{blue}{+28.93}& \textcolor{blue}{+4.41}&\textcolor{blue}{+7.23}&\textcolor{red}{-1.03}\\
         \hline
    \end{tabular}
    }
    \label{tab:downstream_seg}
\end{table}

\subsubsection{Monocular Depth Estimation}
Depth estimation, which involves determining scene depth from a single image, is a crucial component in various applications, including autonomous navigation, 3D reconstruction, and augmented reality. However, most frame-based depth estimation methods~\cite{xian2018monocular,ranftl2020towards,li2018megadepth} encounter challenges in nighttime or low-light conditions due to significantly degraded image quality.

In this section, we use our restoration model to help off-the-shelf frame-based models better adapt to low-light conditions in a two-stage process. We compare the performance of the state-of-the-art monocular depth estimation framework, MiDaS~\cite{ranftl2020towards}, using the original low-light input versus the enhanced input from our EvLight++.

\begin{figure*}[!]
    \centering
    \includegraphics[width=1\linewidth]{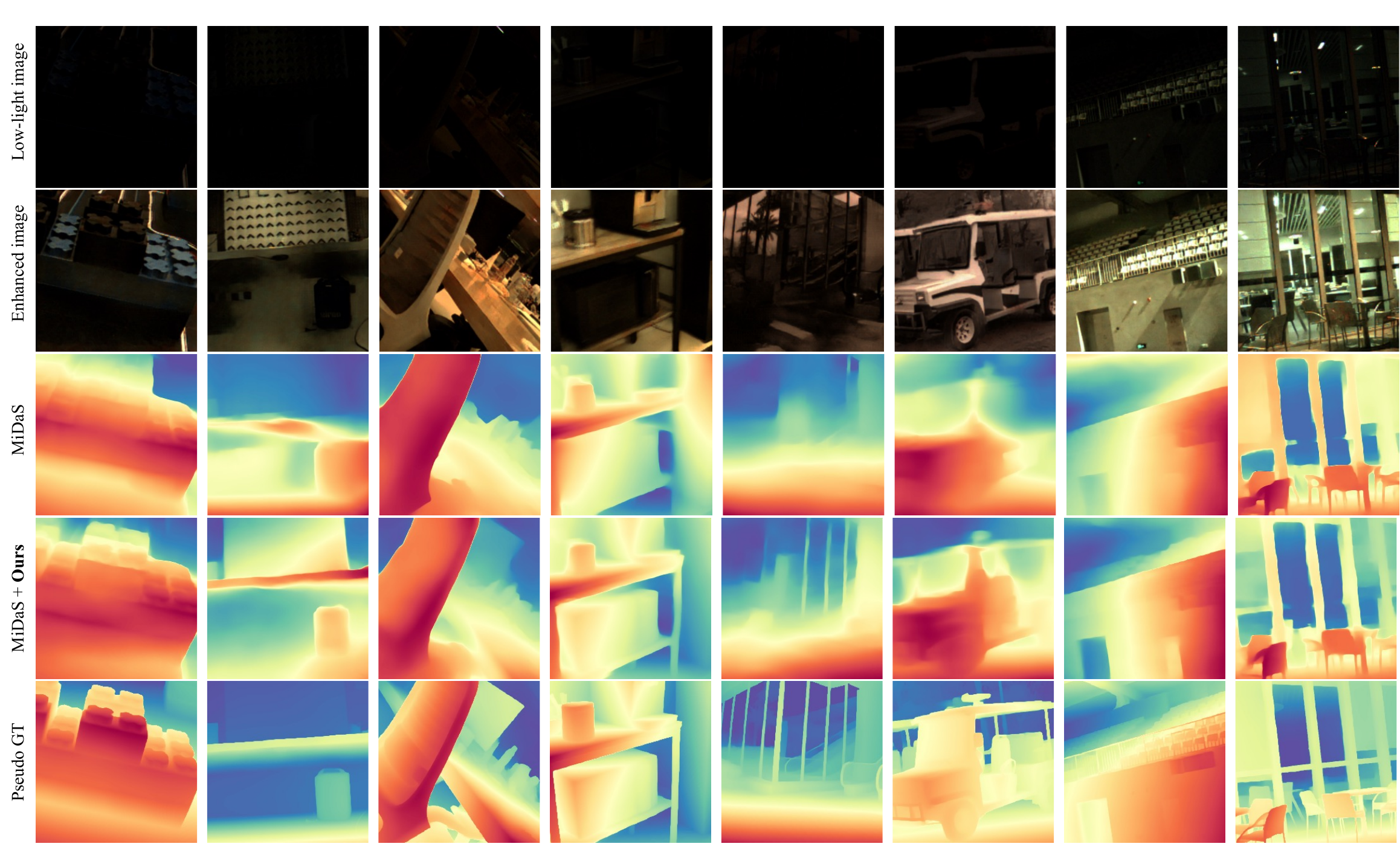}
    \caption{Qualitative comparison of monocular depth estimation result for low-light scenes. The baseline method (\ie MiDaS~\cite{ranftl2020towards}) use the low-light image as input, while ours (\ie MiDaS~\cite{ranftl2020towards}+Ours) uses the enhanced image as input.}
    \label{fig:downstream_depth}
\end{figure*}

To assess the structural accuracy of the predicted depth maps, we calculate the absolute relative error (AbsRel: $|d^*-d|/d$), $\delta_1$ (the percentage of $\max(d^*/d,d/d^*)<1.25$), RMSE, and log10 metrics, using the ground truth from our dataset. The quantitative results in Tab.~\ref{tab:downstream_depth} demonstrate improved performance across all metrics and in both outdoor and indoor scenes when incorporating our method, highlighting its effectiveness. Fig.~\ref{fig:downstream_depth} also provides a qualitative comparison, showing that MiDaS+Ours achieves clearer structures and a closer resemblance to the ground truth depth map, further emphasizing the utility of our method for downstream vision tasks in low-light scenarios.

\begin{table}[!]
    \centering
    \caption{Quantitative results of monocular depth estimation on our SDE dataset. The annotations on the aligned normal-light videos are used as ground truth for calculating evaluation metrics.}
    \setlength{\tabcolsep}{1pt}
    \resizebox{1\linewidth
    }{!}{
    \begin{tabular}{lcccccccc}
    \hline
    \multirow{2}{*}{Method} & \multicolumn{4}{c}{SDE-in}& \multicolumn{4}{c}{SDE-out} \\
     \cmidrule(r){2-5} \cmidrule(r){6-9}
         &  $\delta_1\uparrow$ &  AbsRel$\downarrow$  &log10$\downarrow$ &RMSE$\downarrow$ &  $\delta_1\uparrow$	 &  AbsRel$\downarrow$  &log10$\downarrow$ &RMSE$\downarrow$ \\
         \hline
         MiDaS~\cite{ranftl2020towards}& 0.6416&  0.5471&  0.1322&  0.0995&  0.6194&   0.7222&  0.1424&  0.0991 \\
         MiDaS~\cite{ranftl2020towards}+\textbf{Ours}& 0.6501& 0.4782&  0.1289&  0.0924&   0.6565&   0.6481& 0.1328&  0.0892 \\
         \hline
        \textit{Improve (\%)} & \textcolor{blue}{+1.32} & \textcolor{blue}{+12.59} &\textcolor{blue}{+2.50} &\textcolor{blue}{+7.14} &\textcolor{blue}{+5.99} &\textcolor{blue}{+10.26} &\textcolor{blue}{+6.74} &\textcolor{blue}{+9.99} \\
         \hline
    \end{tabular}
    }
    \label{tab:downstream_depth}
\end{table}

In conclusion, our method and dataset effectively leverage the advantages of event cameras in low-light conditions to facilitate more robust performance in downstream tasks. This highlights the potential of our method to enhance computer vision applications. Moreover, the abundant and precise labels provide a comprehensive benchmark.

\section{Conclusion}


To address the scarcity of paired event data for real-world scenarios, this work presents a large-scale real-world event-image dataset, termed SDE, comprising paired low-light and normal-light videos. To ensure spatial alignment for non-linear motion, we constructed a robotic system equipped with an event camera to capture paired sequences following a predefined identical trajectory. For precise temporal alignment, we introduced a novel matching alignment strategy that minimizes errors with remarkable precision. Based on the high-quality paired real data, we provide abundant and precise labels with well-designed annotation pipelines, facilitating comprehensive evaluation for downstream tasks.
Building on this dataset, we developed EvLight++, a framework that leverages low-light images and HDR event data for robust video enhancement. Our model uses an SNR-guided selection mechanism to adaptively fuse event and image features, both holistically and regionally, ensuring superior performance. Additionally, we incorporated a recurrent module and temporal loss to exploit temporal information and reduce illumination flicker.

\noindent\textbf{Limitations and Future Work:}
Due to the inherent limitations of DAVIS346 event cameras, RGB images in our SDE dataset may exhibit partial chromatic aberrations and the moiré pattern.
In the future, we will improve our hardware system to enable synchronous triggering of robots and event cameras, thereby significantly reducing labor costs associated with repetitive collection.
\ifCLASSOPTIONcaptionsoff
  \newpage
\fi



%
\clearpage
{
    \small
    \bibliographystyle{IEEEtran}
    \bibliography{main}
}




%
\clearpage
\begin{IEEEbiography}[{\includegraphics[width=1in,height=1.25in,clip,keepaspectratio]{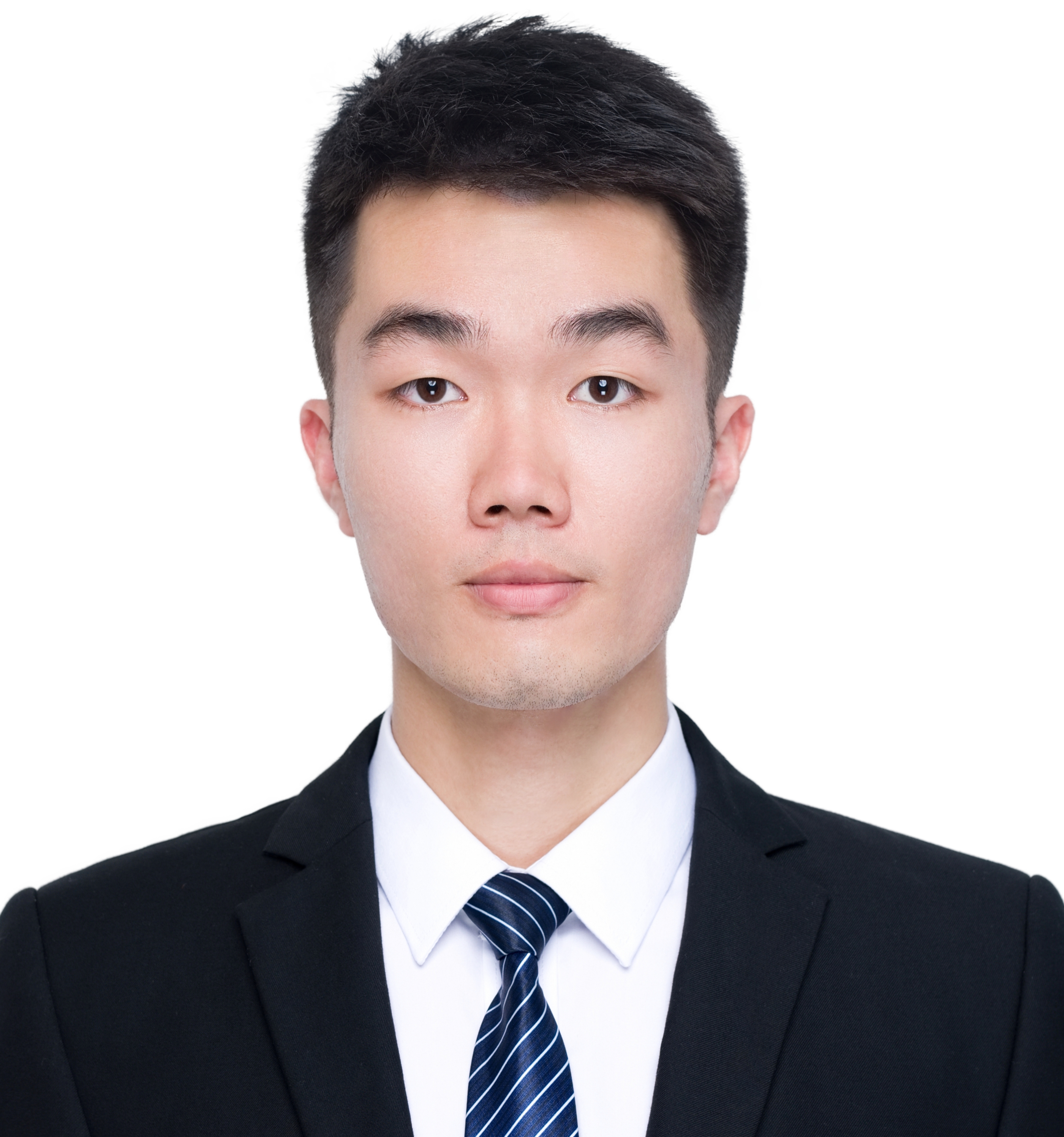}}]
{Kanghao Chen} is currently pursuing a Ph.D. degree in artificial intelligence at the Hong Kong University of Science and Technology (Guangzhou). He received his B.S. degree and M.S. degree in computer science from Jinan University, Sun Yat-Sen University, Guangzhou, China in 2020 and 2022. His research interests include event cameras, 3D reconstruction, and low-level vision. He serves as reviewer for major international conferences, including CVPR, ICCV, NeurIPS.
\end{IEEEbiography}

\begin{IEEEbiography}
[{\includegraphics[width=1in,height=1.25in,clip,keepaspectratio]{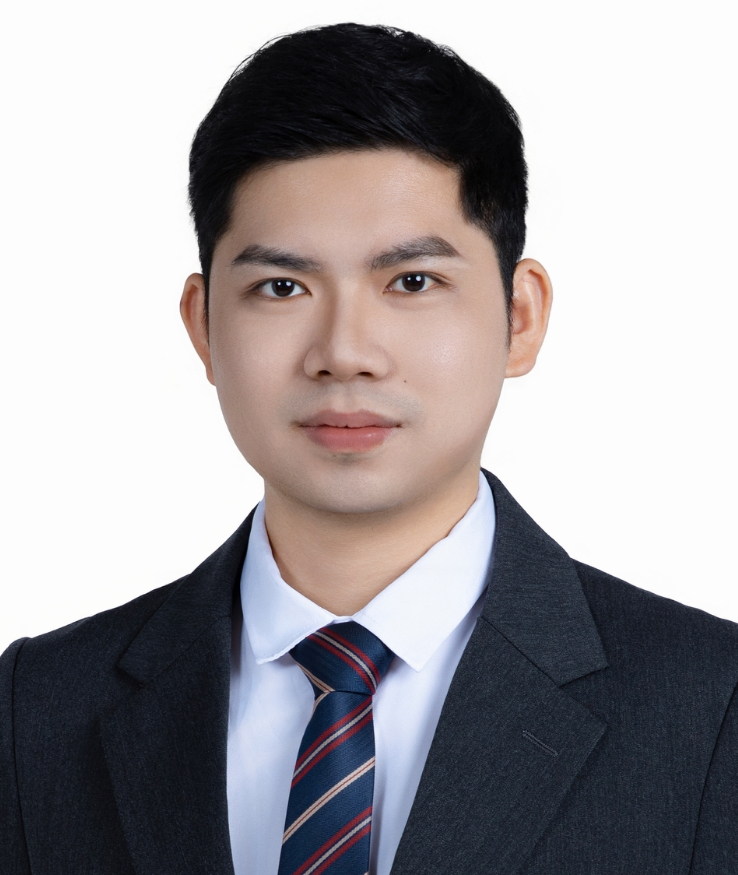}}]
{Guoqiang Liang} is currently studying at Hong Kong University of Science and Technology (Guangzhou). He received his B.S degree in automation from Harbin Institute of Technology (Shenzhen) in 2018. His research areas include diffusion model, image restoration and etc.
\end{IEEEbiography}

\begin{IEEEbiography}
[{\includegraphics[width=1in,height=1.25in,clip,keepaspectratio]{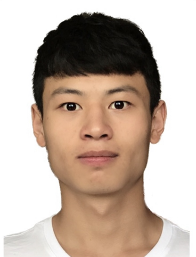}}]
{Hangyu Li} is currently studying at Hong Kong University of Science and Technology (Guangzhou). He received his B.S. degree in Intelligent science and technology from Hebei University of Technology in 2019. His research areas include generative models (ODE, SDE), 3D generation, video generation and etc.
\end{IEEEbiography}

\begin{IEEEbiography}
[{\includegraphics[width=1in,height=1.25in,clip,keepaspectratio]{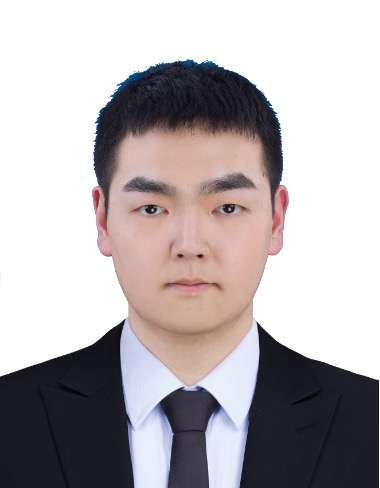}}]
{Yunfan Lu} received the B.S. degree in Computer Science from Nanjing University of Science and Technology and the M.S. degree in Computer Science from the University of Chinese Academy of Sciences. He is currently a final-year Ph.D. student at the Hong Kong University of Science and Technology (Guangzhou). His research interests include computational imaging, knowledge mining from image and video data, and event camera technology. He has served as a reviewer for major international conferences and journals, including CVPR, ECCV, T-PAMI, and IJCV.
\end{IEEEbiography}

\begin{IEEEbiography}[{\includegraphics[width=1in,height=1.25in,clip,keepaspectratio]{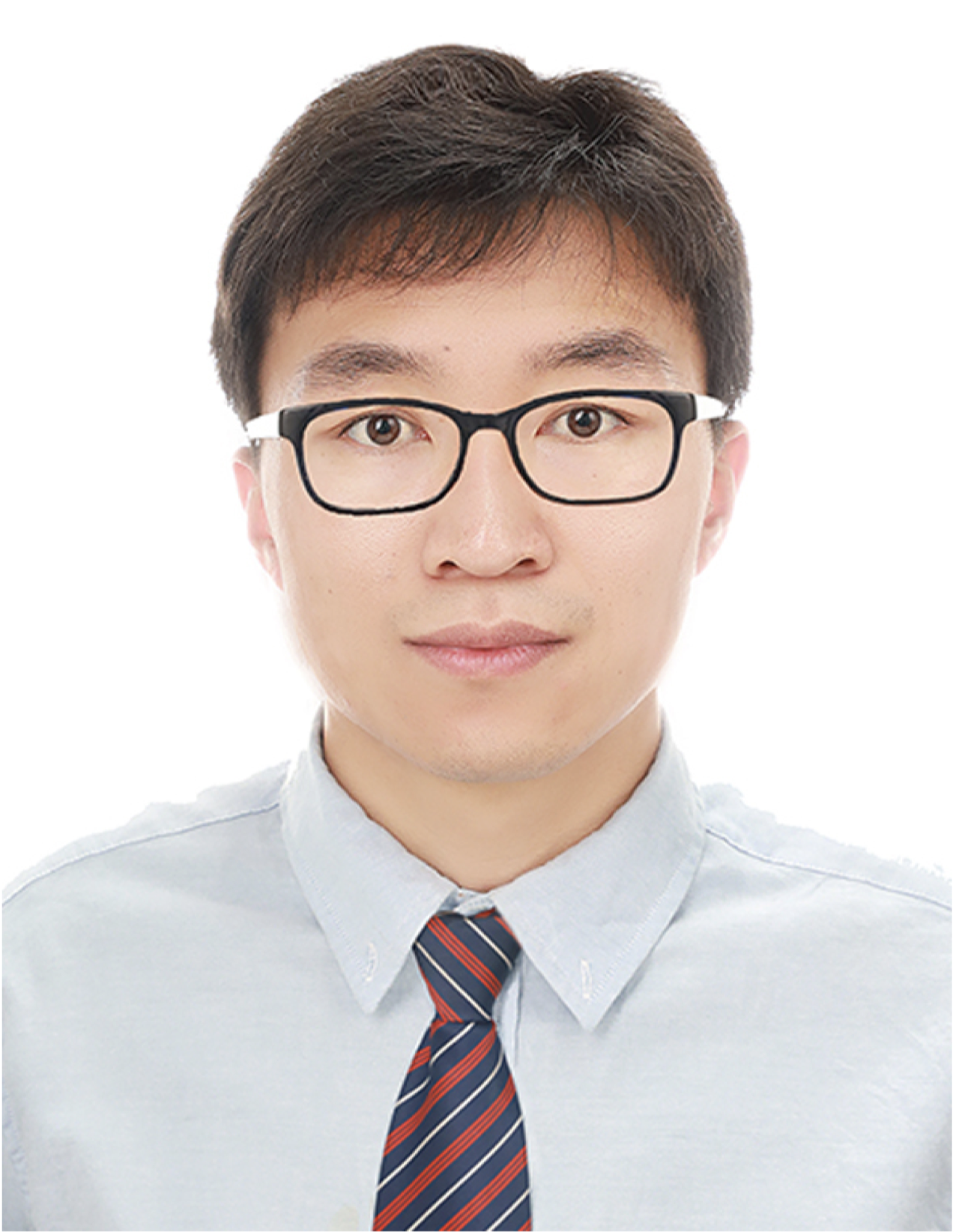}}] 
{Lin Wang} (IEEE Member) is an assistant professor in the AI Thrust, HKUST-GZ, HKUST FYTRI, and an affiliate assistant professor in the Dept. of CSE, HKUST. He did his Postdoc at the Korea Advanced Institute of Science and Technology (KAIST). He got his Ph.D. (with honors) and M.S. from KAIST, Korea. He had rich cross-disciplinary research experience, covering mechanical, industrial, and computer engineering. His research interests lie in computer and robotic vision, machine learning, intelligent systems (XR, vision for HCI), etc. 
\vspace{-50pt}
\end{IEEEbiography}








\end{document}


\maketitle

\begin{abstract}
Due to the lack of space in the main paper, we provide more details of the proposed method and experimental results as well as the dataset in the supplementary material.
Specifically, we provide the collection of the dataset in Section 1. 
Section 3 provides additional analysis.
Section 2 describes more experimental and results.
\end{abstract}



\begin{table}[]
\centering
\resizebox{0.45\textwidth}{!}{
\begin{tabular}{lllll}
\hline
\multicolumn{2}{l}{Method}                                        & PSNR & FLOPs (G) & \#Params (M) \\ \hline
Event Only                   & E2VID+ (ECCV'20)                   &15.19      &27.99      &10.71          \\ \hline
\multirow{4}{*}{Image Only}  & SNR-Net (CVPR'22)                  &20.05      &26.35    & 4.01           \\
                             & Uformer (CVPR'22)                  &21.09      &12.00    & 5.29           \\
                             & LLFlow-L-SKF (CVPR'23)             &20.92      &409.50     & 39.91          \\
                             & Retinexformer (ICCV'23)            &21.30      &15.57    & 1.61           \\ \hline
\multirow{4}{*}{Image+Event} & ELIE (TMM'23)                      &19.98      &440.32   & 33.36         \\
                             & Esl-Net (ECCV'20                   &21.25      &560.94   & 0.56           \\
                             & Liu \etal (AAAI'23)                &21.79      &44.71    & 47.06          \\
                             & Ours                               &22.44      &180.90   & 22.73          \\ \hline
\end{tabular}
}
\end{table}


\section{Dataset Collection}
\label{sec:dataset_collection}
\begin{figure}[t]
\centering
    \includegraphics[width=1\linewidth]{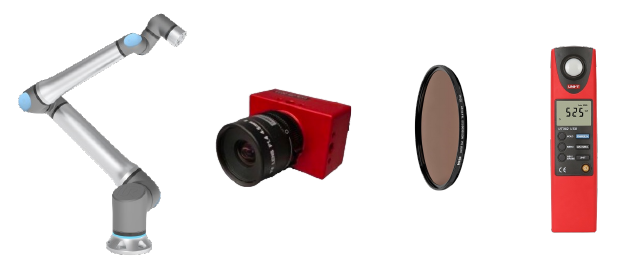}
    \caption{The devices used for collecting our dataset. From the left to right is the Universal Robot UR5e, an ND8 filter, the DAVIS 346 event camera, and an illuminance meter. 
    }
    \vspace{-10pt}
    \label{fig:dataset-device}
\centering
\end{figure}

\begin{figure*}[t]
\centering
    \includegraphics[width=1\linewidth]{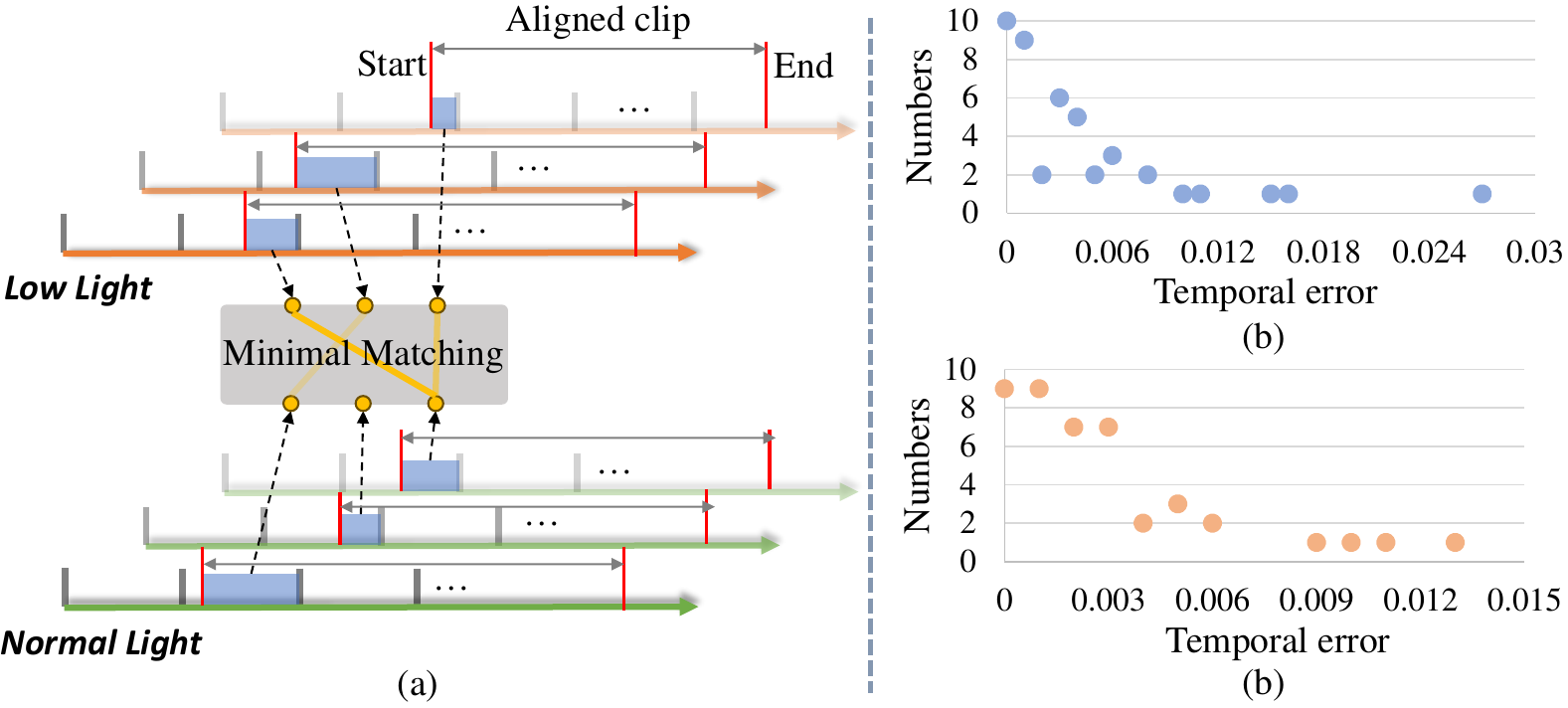}
    \caption{The devices used for collecting our dataset. From the left to right is the Universal Robot UR5e, an ND8 filter, the DAVIS 346 event camera, and an illuminance meter. 
    }
    \vspace{-10pt}
    \label{fig:dataset-device}
\centering
\end{figure*}

\begin{figure*}[t]
\centering
    \includegraphics[width=1\linewidth]{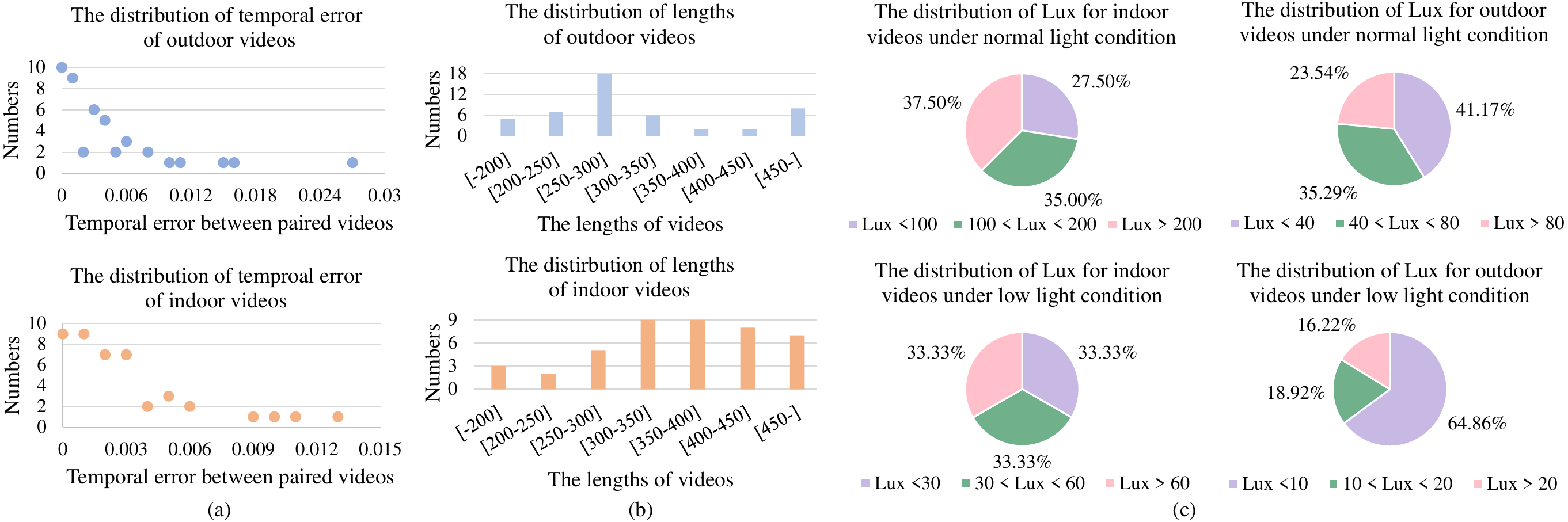}
    \caption{The devices used for collecting our dataset. From the left to right is the Universal Robot UR5e, an ND8 filter, the DAVIS 346 event camera, and an illuminance meter. 
    }
    \vspace{-10pt}
    \label{fig:dataset-device}
\centering
\end{figure*}

\begin{figure*}[t]
\centering
    \includegraphics[width=1\linewidth]{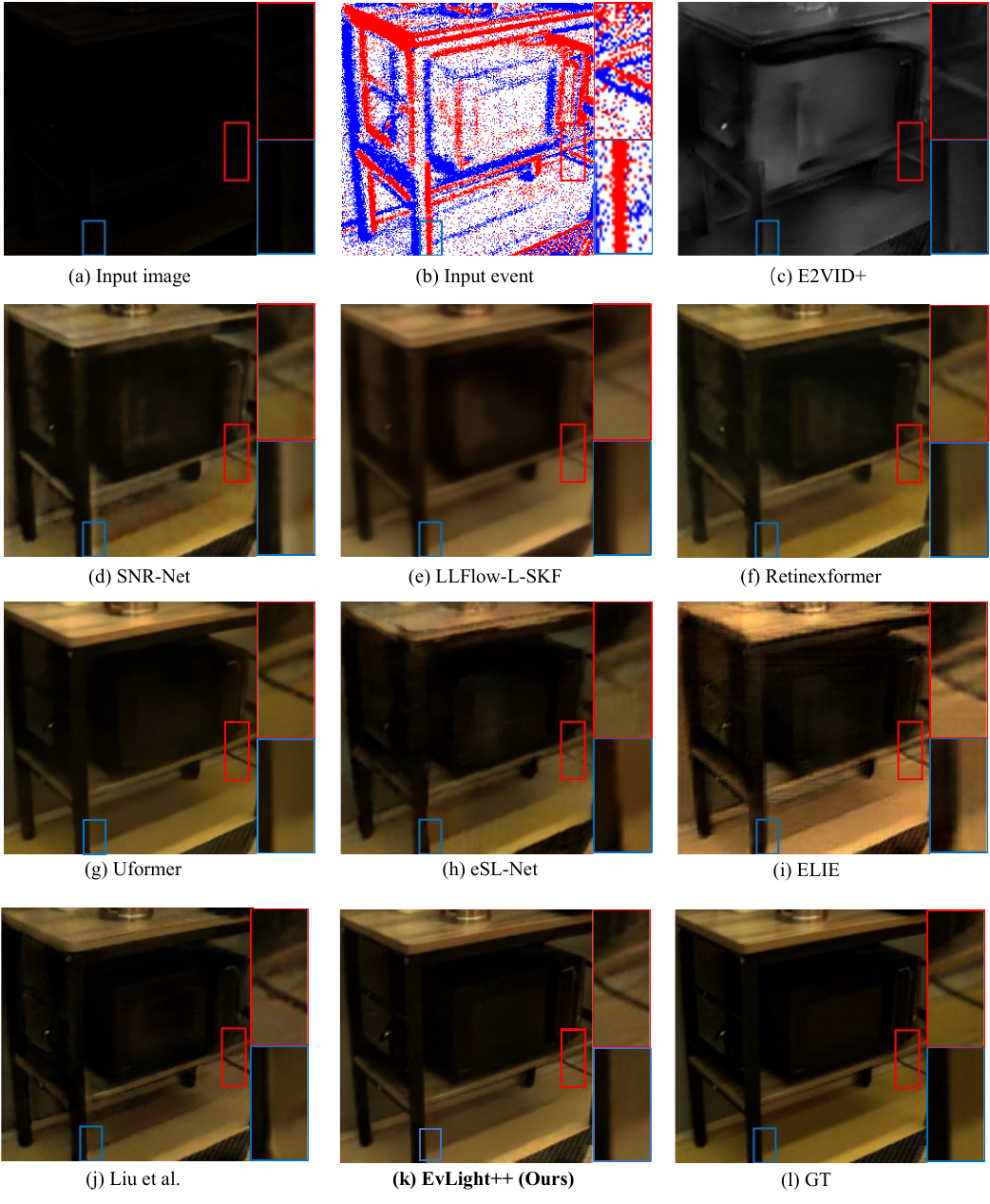}
    \caption{The devices used for collecting our dataset. From the left to right is the Universal Robot UR5e, an ND8 filter, the DAVIS 346 event camera, and an illuminance meter. 
    }
    \vspace{-10pt}
    \label{fig:dataset-device}
\centering
\end{figure*}

\begin{figure*}[t]
\centering
    \includegraphics[width=1\linewidth]{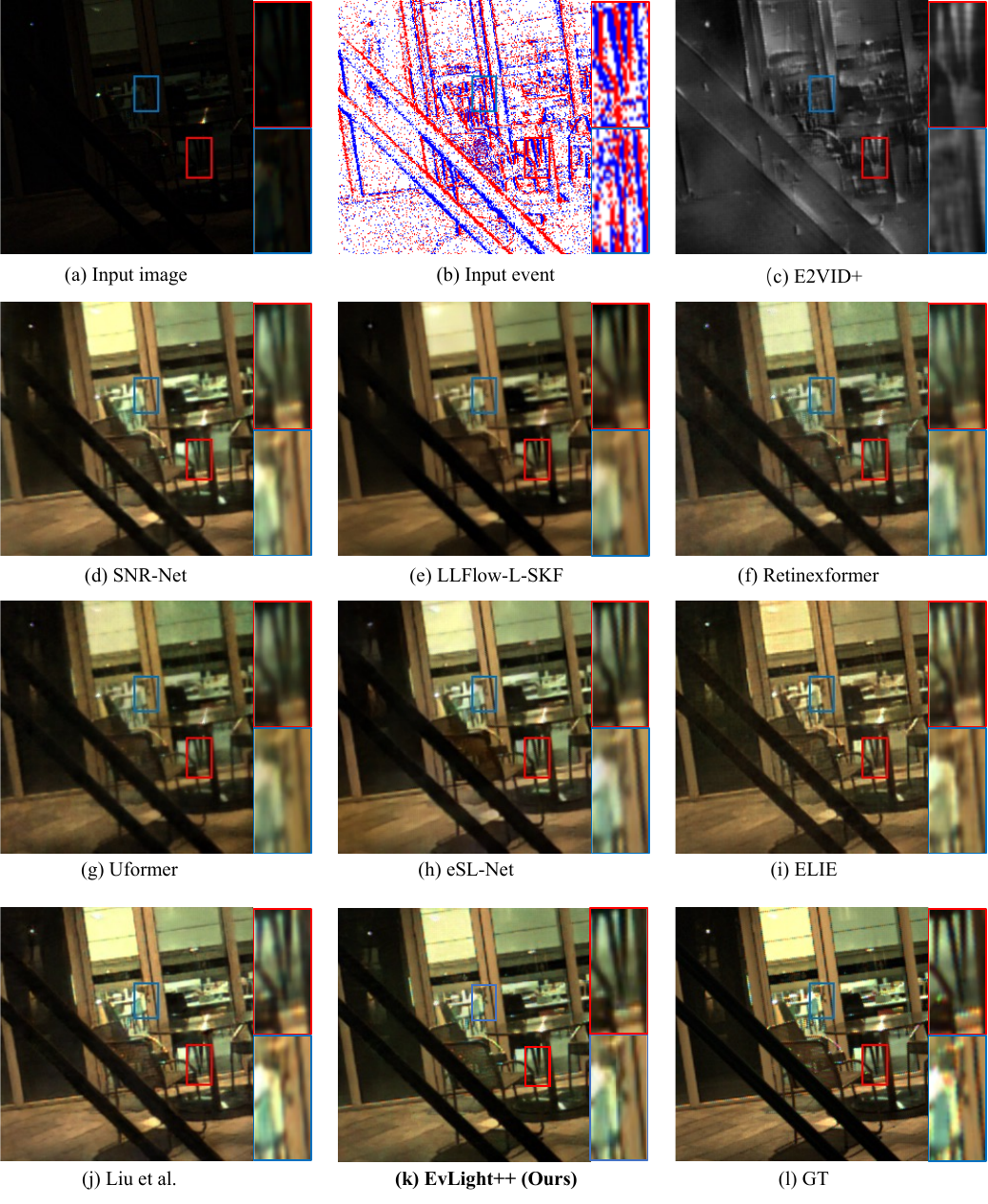}
    \caption{The devices used for collecting our dataset. From the left to right is the Universal Robot UR5e, an ND8 filter, the DAVIS 346 event camera, and an illuminance meter. 
    }
    \vspace{-10pt}
    \label{fig:dataset-device}
\centering
\end{figure*}

\begin{figure*}[t]
\centering
    \includegraphics[width=1\linewidth]{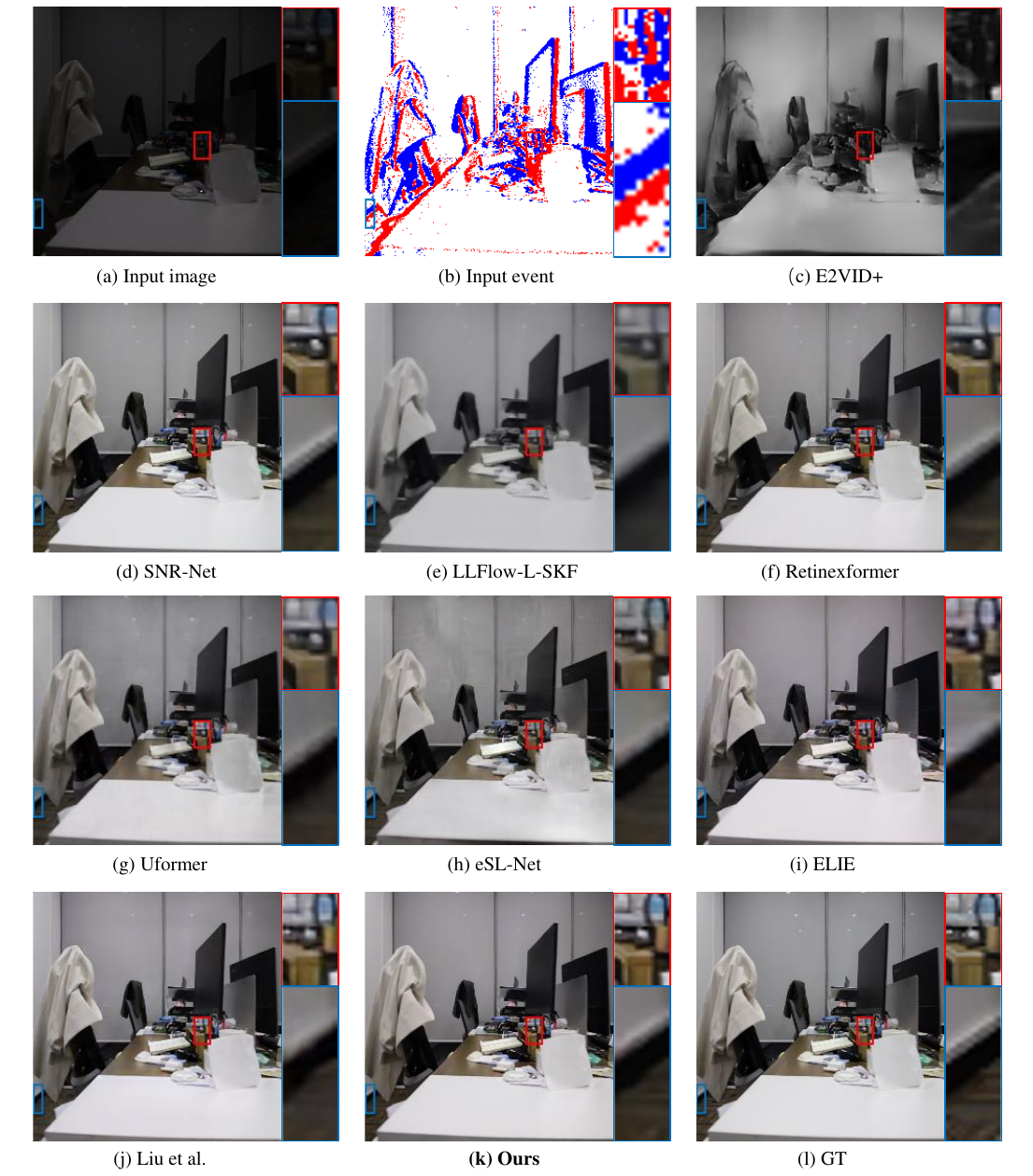}
    \caption{The devices used for collecting our dataset. From the left to right is the Universal Robot UR5e, an ND8 filter, the DAVIS 346 event camera, and an illuminance meter. 
    }
    \vspace{-10pt}
    \label{fig:dataset-device}
\centering
\end{figure*}

\begin{figure*}[t]
\centering
    \includegraphics[width=1\linewidth]{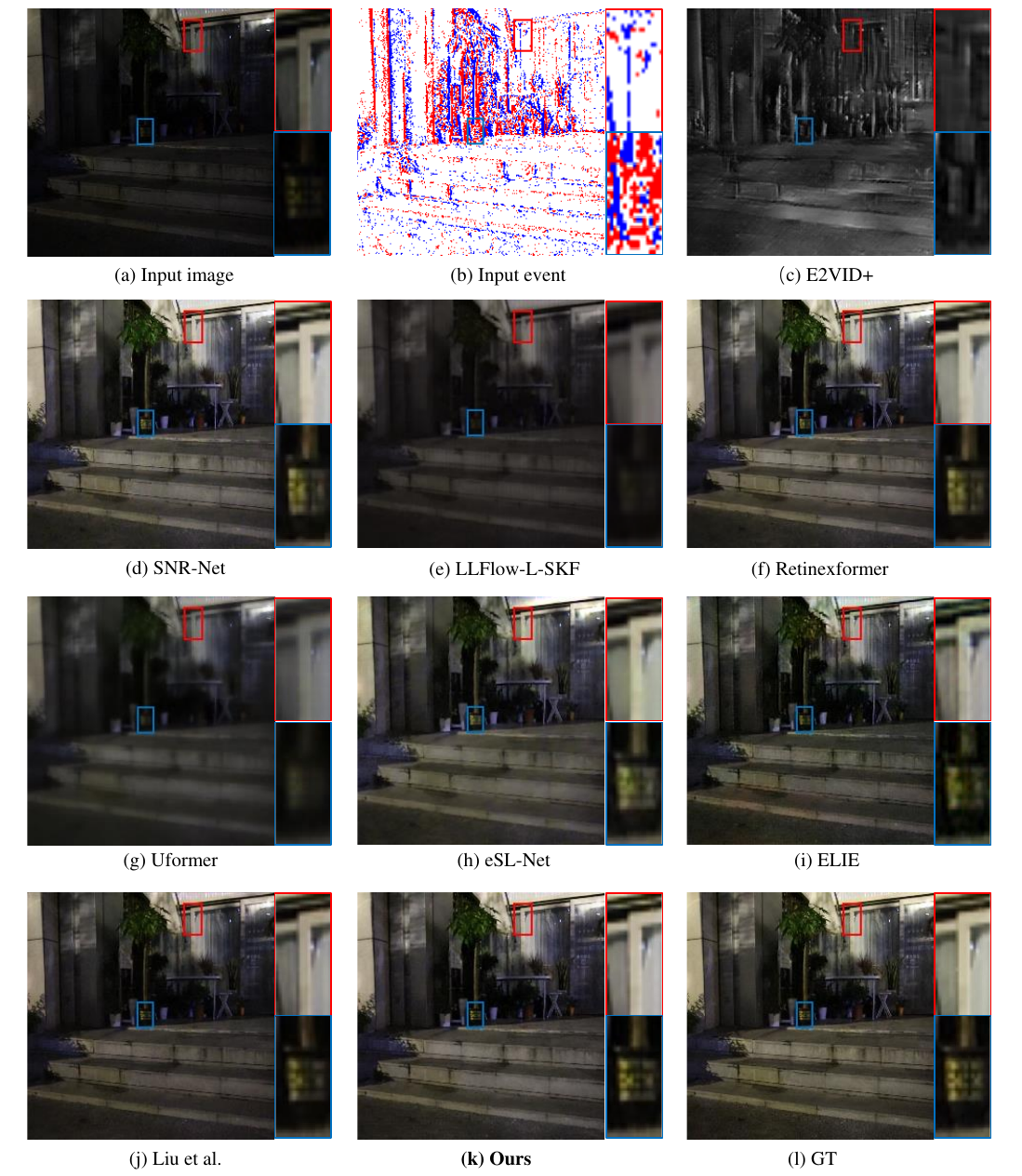}
    \caption{The devices used for collecting our dataset. From the left to right is the Universal Robot UR5e, an ND8 filter, the DAVIS 346 event camera, and an illuminance meter. 
    }
    \vspace{-10pt}
    \label{fig:dataset-device}
\centering
\end{figure*}

\begin{figure*}[t]
\centering
    \includegraphics[width=1\linewidth]{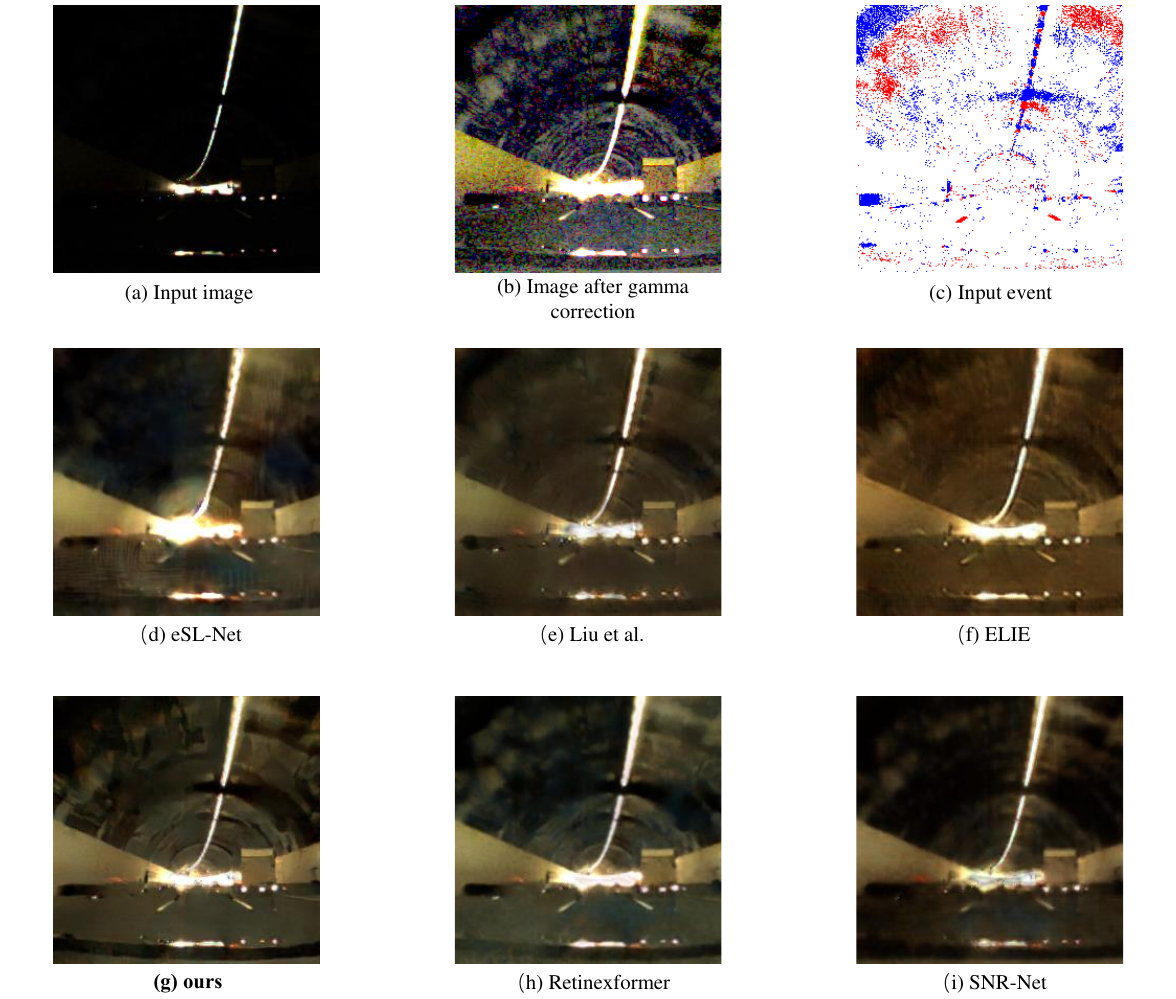}
    \caption{Generalization results in the driving tunnel sun sequence captured in CED~\cite{scheerlinck2019ced} dataset.
    }
    \vspace{-10pt}
    \label{fig:dataset-device}
\centering
\end{figure*}

\begin{figure*}[t]
\centering
    \includegraphics[width=1\linewidth]{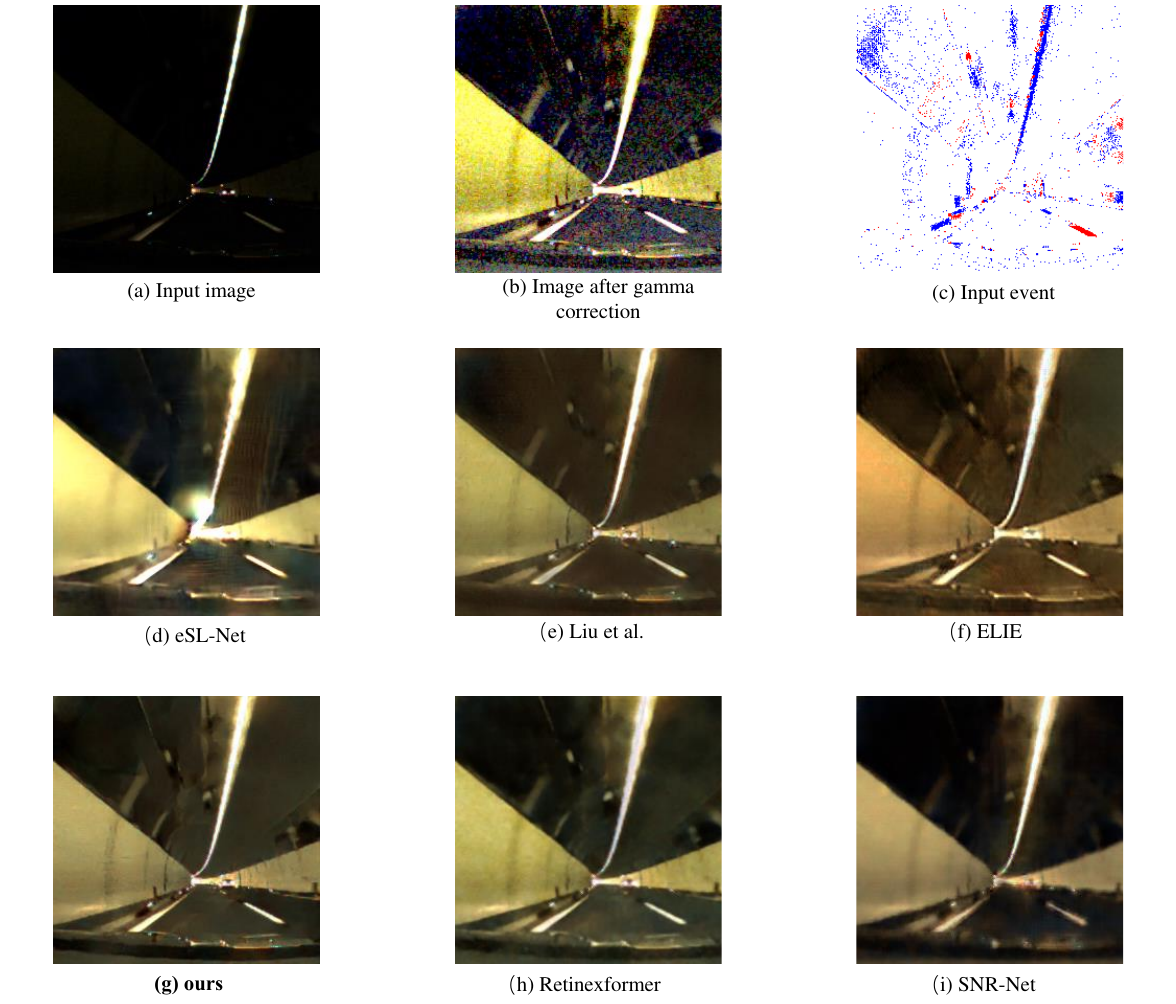}
    \caption{Generalization results in the driving tunnel sequence captured in CED~\cite{scheerlinck2019ced} dataset.
    }
    \vspace{-10pt}
    \label{fig:dataset-device}
\centering
\end{figure*}

ApsenTek created the event camera that we used to get the real data. \footnote{https://www.alpsentek.com/product}.
The active pixel sensor (APS) and dynamic vision sensor (DVS) of this camera share an optical system.
APS generate RGB signals.
DVS generate event signals.
In addition, APS and DVS are aligned in physical, as shown in Fig.\ref{fig:sensor}.
Both event signals and RGB signals can be spatially positioned in alignment using this method.
APS are quad-bayer pattern and enjoy a $3264 \times 2448$ resolution.
DVS enjoy a $1632 \times 1224$ spatial resolution and $300Hz$ temporal resolution.
DVS has a worse spatial resolution than APS by 50\%.
While DVS pixels capture the same color as APS.
For instance, in Fig.\ref{fig:sensor}, the pixel in the top left corner only detects green light, whether it is an APS or DVS pixel.
Because APS and DVS share an optical system.

APS uses a rolling shutter exposure method, as shown in Fig.\ref{fig:sensor-pattern}.
In practice, the interval from the exposure start time $T_s$ to the exposure end time $T_e$ is larger than the exposure time $T_{exp}$, as shown in Fig.\ref{fig:sensor-pattern} (a).
We select all events with timestamps between $T_s$ and $T_e + T_{exp}$ as events aligned to the intensity image.

The exposure mode of DVS is output every $1/400s$, which can be approximated as global exposure.
The output's polarity is determined by the brightness values at the present and the preceding instant.
After our test $1/400s$ as DVS temporal resolution can achieve a balance between temporal resolution and event quality.

\section{More Experiments and Visualization}
\label{sec:more_experiments_and_visualization}

More visualization results are shown in Fig.\ref{fig:01-CED8x-SuM} and Fig.\ref{fig:01-CED8x-SuM-2} Our method can recover more details and reduce artifacts.